\RequirePackage{fix-cm}
\documentclass[twocolumn]{svjour3}         

\smartqed  

\usepackage{epsfig}
\usepackage{microtype}
\usepackage{hyperref}
\usepackage{url}            
\usepackage{bm}
\usepackage{amssymb,amsfonts,dsfont,mathrsfs}
\usepackage{listings}
\usepackage{booktabs,multirow,adjustbox}
\usepackage{subfig}
\usepackage{amsmath}
\usepackage{color}
\usepackage{xcolor}
\usepackage{colortbl}
\usepackage{mathtools}
\usepackage{pifont}
\usepackage{float}
\usepackage{verbatim}
\usepackage{lipsum}
\usepackage{natbib}
\usepackage{array}
\usepackage{graphicx}
\usepackage{times}
\definecolor{uared}{rgb}{0.85, 0.0, 0.3}

\newcommand{\FID}{\textbf{FID$\downarrow$}}  
\newcommand{\MSE}{\textbf{MSE$\downarrow$}}  
\newcommand{\PSNR}{\textbf{PSNR$\uparrow$}}  
\newcommand{\SSIM}{\textbf{SSIM$\uparrow$}}  
\newcommand{\SWD}{\textbf{SWD$\downarrow$}}  

\newcommand{\redz}{\textcolor[rgb]{1,0,0}{\bm z}}
\newcommand{\bluey}{\textcolor[rgb]{0,0,1}{\bm y}}

\begin{document}

\title{Disentangled Inference for GANs with Latently Invertible Autoencoder}


\author{Jiapeng Zhu\thanks{* denotes equal contribution. Part of the work was done when Zhu was an intern at Xiaomi AI Lab.}\textsuperscript{$\ast$1,2} \and
        Deli Zhao\textsuperscript{$\ast$1} \and
        Bo Zhang\textsuperscript{1} \and
        Bolei Zhou\textsuperscript{2}
}

\institute{
  \textsuperscript{1}Xiaomi AI Lab, China. \\
  \textsuperscript{2}Department of Information Engineering, The Chinese University of Hong Kong, Hong Kong, China.
}

\date{}

\maketitle

\begin{abstract}
Generative Adversarial Networks (GANs) can synthesize more and more realistic images. However, one fundamental issue hinders their practical applications: the incapability of encoding real samples in the latent space. Many semantic image editing applications rely on inverting the given image into the latent space and then manipulating inverted code. 
One possible solution is to learn an encoder for GAN via Variational Auto-Encoder (VAE). However, the entanglement of the latent space poses a major challenge for learning the encoder. To tackle the challenge and enable inference in GANs, we propose a novel method named Latently Invertible Autoencoder (LIA). In LIA, an invertible network and its inverse mapping are symmetrically embedded in the latent space of an autoencoder. The decoder of LIA is first trained as a standard GAN with the invertible network, and then the encoder is learned from a disentangled autoencoder by detaching the invertible network from LIA. It thus avoids the entanglement problem caused by the latent space.  Extensive experiments on the FFHQ face dataset and three LSUN datasets validate the effectiveness of LIA for the image inversion and its applications. Code and models are available at \url{https://github.com/genforce/lia}.

\end{abstract}

\section{Introduction}
\label{intro}
Deep generative models play more and more important roles in cracking challenges in computer vision and other fields, such as high-resolution image generation~\citep{pix2pix,cycleGAN,PGGAN,StyleGAN18,BigGAN}, text-to-speech transformation~\citep{waveNet16,waveNet17}, information retrieval~\citep{IRGAN}, 3D rendering~\citep{Jiajun16,sceneRendering18}, and signal-to-image acquisition~\citep{signal-to-image18}. In particular, Generative Adversarial Network (GAN)~\citep{GAN} exhibits an extraordinary capability of learning the distribution of high-dimensional imagery data. For example, it is now very difficult to distinguish the face images synthesized by StyleGAN algorithms~\citep{StyleGAN18,StyleGAN2019} from real face images.

However, a critical limitation for the vanilla GAN is its incapability for encoding real samples. Namely, we cannot infer the latent variable $\bm z$ corresponding to a given \real sample $\bm x$ such that the image can be faithfully reconstructed from $\bm z$ by the GAN generator. The problem is critical because many applications usually depend on manipulating the latent code such as the domain adaptation~\citep{Sankaranarayanan2017domain}, the data augmentation~\citep{Antoniou2017augmentation}, and the image editing~\citep{image2stylegan,shen2020interpreting,Zhu2020Indomain}. 

The existing approaches for addressing this issue fall into three categories, as summarized in Table~\ref{tab:inference-methods}. The common approach is called GAN inversion that is based on the optimization of Mean Squared Error (MSE) between generated samples and real samples~\citep{DCGAN16,BEGAN17,image2stylegan}. This type of algorithms mainly has two drawbacks: the sensitivity to the initialization of $\bm z$ and the slow iterative optimization process. The second category is named as adversarial inference~\citep{ALI2017,BiGAN2017}, which uses another GAN to infer $\bm z$ as latent variables in a framework of dual GANs. Adversarial inference generally aims to learn high-level discriminative features used for classification rather than faithfully reconstruct samples~\citep{BigBiGAN2019}. The third approach for GAN inference is to finetune an encoder via the principle of the VAE algorithm, given a pretrained GAN model~\citep{inverseGenerator17} or to learn jointly the VAE and GAN architectures in an end-to-end manner~\citep{VAEGAN}. Combining VAE and GAN together with shared decoder (the VAE/GAN method) sounds like an elegant solution to GAN inference. However, the problem of integrating VAE and GAN is that the reconstruction precision is usually worse than that of using VAE alone and the quality of reconstructed samples is inferior to that of generated samples from sampling pretrained GANs. We analyze this problem in detail in section~\ref{sec:experiment}.

\setlength{\tabcolsep}{10pt}
\begin{table}[t]
\centering
\caption{Different approaches for enabling inference in GANs. $f$ and $g$ denote the encoder and the decoder (or generator in GAN notation) in each method, respectively. $c$ is the discriminator for GAN. $\bm z$ follows a probabilistic prior $\bm z \sim p(\bm z)$ and $\bm y = \phi^{-1}(\bm z)$ where $\phi$ is an invertible network. }
  \begin{tabular}{  l | c  }
    \hline
    Method    &  Formulation\\
     \hline 
    GAN inversion &   ${\bm z}^{\ast} = \arg\min_{\bm z} \|g(\bm z) -\bm x\|$ \\
    \hline
    Adversarial inference &  $\bm x \overset{f}{\mapsto} \tilde{\bm z}, ~ \{
  \tilde{\bm z}, \bm z\}
\overset{c}{\mapsto} 0/1 $  \\
     \hline
 \multirow{2}{*}{VAE/GAN} & 
$ \bm x \overset{f}{\mapsto} \redz \overset{g}{\mapsto} \tilde{\bm x}, ~ \{
  \tilde{\bm x}, \bm x\}
\overset{c}{\mapsto} 0/1 $\\
  & ~~~~~~~~~~~~~~~~~~~~~~~$\max \log p(\bm x)$\\
 \hline
 \multirow{2}{*}{LIA/GAN} & 
$ \bm x \overset{f}{\mapsto} \bluey \overset{g}{\mapsto} \tilde{\bm x}, ~ \{
  \tilde{\bm x}, \bm x\}
\overset{c}{\mapsto} 0/1 $\\
   & ~~~~~~~~~~~~~~~~~~~~~~$\min\|\bm x-\tilde{\bm x}\|$\\
 \hline
  \end{tabular}
  \label{tab:inference-methods}
\end{table}

In this work, we show that the disentanglement of the latent space plays a crucial role in learning a high-quality encoder for GAN. The entanglement of the $\bm z$-space is the underlying reason why the combination of VAE/GAN don't work well. Inspired by the mapping network design in StyleGAN~\citep{StyleGAN18}, we explore the property of the intermediate latent space (the $\bm y$-space in this paper) that is the output of the mapping network and reveal its disentanglement property. 
Based on the disentanglement property, we develop a new method called Latently Invertible Autoencoder (LIA). LIA utilizes an invertible network to bridge the encoder and the decoder in a symmetric manner in the latent space, and then follows two-stage training scheme to enable accurate inference and reconstruction for GANs.
We summarize the contributions as follows:
\begin{itemize}
    \item We analyze the degree of disentanglement in the latent space ($\bm z$-space) and in the intermediate latent space ($\bm y$-space) respectively, and show that the entanglement in GAN's latent space is the key reason why the previous VAE-based encoder methods don't work well. Based on this analysis, we propose the LIA architecture and the corresponding two-stage training scheme. 
    \item The symmetric design of the invertible network in LIA brings the advantage that the prior distribution can be faithfully embedded in a \textit{disentangled} latent space, significantly easing the inference problem. 
    Besides, the two-stage training scheme decomposes the LIA framework into the vanilla GAN with an invertible network and a standard autoencoder without stochastic variables. Therefore the encoder training can be conducted in the \textit{disentangled latent space} without the stochastic variational inference, overcoming its mode collapse and convergence issues.
    \item We compare LIA with other methods in terms of inference and reconstruction. The experimental results on FFHQ and LSUN datasets show the proposed method achieves superior performance. 
\end{itemize}

\section{Preliminary}
In this section, we introduce the basics of GAN, the inference method with VAE/GAN, and the inversion approach. 
\subsection{Generative Adversarial Network}
GAN uses the adversarial training to learn data distribution from a given dataset $\mathcal{X}$~\citep{GAN}. The framework of GAN consists of a generator $g(\cdot)$ that generates fake images from random vectors $\bm z$, $i.e$. $\tilde{\bm x} = g(\bm z)$, and a discriminator that distinguishes fake images from real ones. The discriminator can be also called the critic~\citep{WGAN17}, thus here we use $c(\cdot)$ to represent the discriminator function. The adversarial training is defined as the following min-max optimization problem,
\begin{align}\label{eq:gan-loss}
g^{\ast}, c^{\ast} &= \arg\min_{g}\max_{c} V(g,c),
\end{align}
where $V(g,c) = \mathbb{E}_{p(\bm x)}[\log c(\bm x)] + \mathbb{E}_{p(\bm z)}[\log (1-c(g(\bm z)))]$ 
and $p(\bm x)$ is the distribution of the real image data and $p(\bm z)$ is the prior distribution for latent variable $\bm z$. 
In practice, the optimization can be solved by alternating update of a minimum problem $\min_{g} V(g,c)$  and a maximum one $\max_{c} V(g,c) $. 
The generator is used to generate realistic images from randomly sampled $\bm z$ after training.

StyleGAN algorithms~\citep{StyleGAN18,StyleGAN2019}  shed new light on high-quality image generation by introducing an intermediate latent space. Compared to previous GANs, StyleGAN learns a mapping network to map $\bm z$ into an intermediate latent space $\bm y$-space and then uses style transfer operator AdaIN \citep{Huang2017AdaIN} to expand $\bm y$ to convolutional layers of the generator. The transformation of the variables in StyleGAN can be written as
\begin{equation}\label{eq:diagramStyleGAN}
\bm z \overset{\varphi}{\longmapsto}  \bm y \overset{g}{\longmapsto} \tilde{\bm x} \leftrightarrow 
\Big \{
\begin{matrix}
  \tilde{\bm x} \\
  \bm x
 \end{matrix} \Big\}
\overset{c}{\longmapsto}  V(g,c), 
\end{equation}
where $\varphi$ denotes the mapping network.

An obvious drawback of the original GAN framework is that we cannot directly obtain a latent code $\bm z$ that can reconstruct a given real sample by the trained generator due to the lack of an encoder. To address this issue, there are two main approaches summarized below.

\subsection{Inference with VAE}
VAE is composed of two dual mappings, $i.e.$ $ \bm x \overset{f}{\mapsto} \bm z \overset{g}{\mapsto} \tilde{\bm x}$, where $f$ is the encoder and $g$ is the decoder. 
Let $p_g(\bm x|\bm z)$ denote the likelihood of generated sample conditioned on the latent code $\bm z$ and $q_f(\bm z|\bm x)$ the posterior distribution. Variational autoencoder optimizes the lower bound of the marginal log-likelihood~\citep{VAE}
\begin{equation}
  \log p_{f,g}(\bm x)  \approx  -\text{KL}[q_f(\bm z|\bm x) || p(\bm z)] + \mathbb{E}_q[\log p_g(\bm x|\bm z)], 
\end{equation}
where ${\text{KL}}[q_f(\bm z|\bm x) || p(\bm z)]$ is the Kullback-Leibler divergence. 
The first term $\text{KL}[q_f(\bm z|\bm x) || p(\bm z)]$ constrains the latent code to the prior via the KL-divergence, and the second term $\mathbb{E}_q[\log p_g(\bm x|\bm z)]$ guarantees the reconstruction accuracy. For a Gaussian distribution $p_g(\bm x|\bm z)$ with diagonal covariance matrix, $\log p_g(\bm x|\bm z)$ reduces to the variance-weighted squared error~\citep{VAE-tutorial16}.

The inference for real images through VAE is done by the encoder to output $\bm z$ in the latent space of GAN, thus the optimization of the encoder $f$ is as follows,
\begin{equation} \label{eq:VAE-opt}
f^{\ast},\hat{c}^{\ast} = \arg\max_{f,c^{\ast}} V(g^{\ast},c^{\ast}) + \alpha_{\text{vae}} \log p_{f,g^{\ast}}(\bm x),
\end{equation}
 where $g^{\ast}$ and $c^\ast$ mean that $g$ and $c$ have been already derived with GAN and $\alpha_{\text{vae}}$ is a hyper-parameter. The issue of incorporating VAE into GAN framework is that the $\bm z$-space is entangled, thus the optimization in Equation~(\ref{eq:VAE-opt}) usually results in sub-optimal solutions. We will explain this issue in more detail in section~\ref{sec:entanglement}.

\begin{figure*}[t]
\begin{center}
\begin{tabular}{c}
    \includegraphics[scale=1]{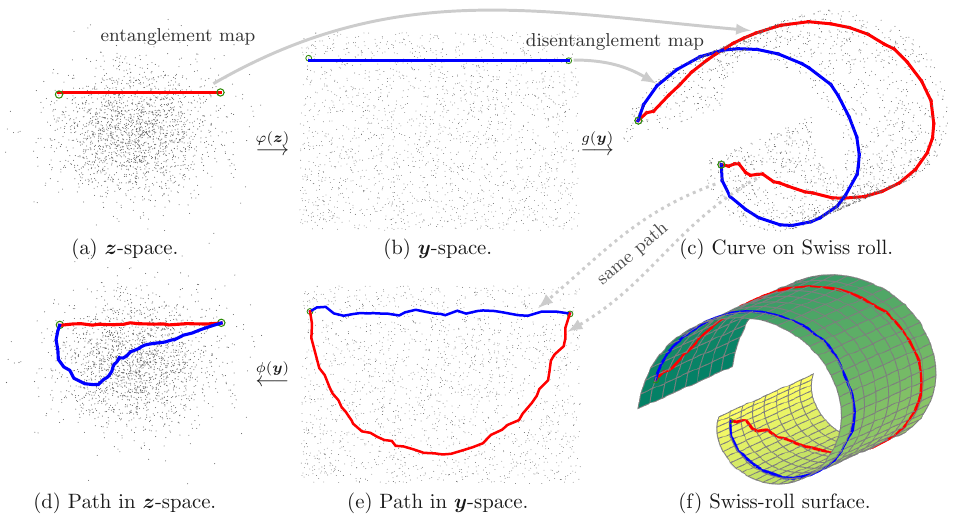}\\
\end{tabular}
\end{center} \vspace{-0.6cm}
  \caption{Illustration of the disentanglement via the Swiss-roll manifold. Swiss roll in (c) is obtained from the roll-shaped functional mapping with coordinates in (b).  So the blue line in (b) corresponds to the shortest blue path (geodesic on Swiss-roll) in (c). The paths in (e) are the same ones in (c). The red path in (c) and the blue path in (d) are manually shaped for comparison. There are no explicit functions for $\varphi$ and $\phi$ here. The shape of the $\bm y$-space cannot be directly computed from the $\bm z$-space in this figure. So they are plotted as illustration.
  }
\label{fig:toy}
\vspace{-0.3cm}
\end{figure*}

\subsection{GAN Inversion}
GAN inversion aims to compute the latent code $\bm z$ by minimizing the squared error between a given real sample $\bm x$ and its reconstruction $\tilde{\bm x} = g(\bm z)$ as
\begin{equation}\label{eq:gan-inversion2}
    {\bm z}^{\ast} = \arg\min_{\bm z} \|\tilde{\bm x} -\bm x\| + \alpha_{\text{vgg}}\sum^l_{i=1} \| F_i(\tilde{\bm x}) - F_i(\bm x) \|.
\end{equation}

The first term in Equation~(\ref{eq:gan-inversion2}) is the pixel distance between the two images and the second term is the perceptual distance~\citep{perceptualLoss2016} in feature space. 
$F_i$ is the feature map of the $l$-th layer in the pretrained VGG network~\citep{VGG}.
Though this method is simple and doesn't need to train an encoder, its limitations lie in two aspects. The first one is that the precision of ${\bm z}^{\ast}$ heavily relies on the initial value ${\bm z}_0$ (see the analysis in section~\ref{exp:finetuning}); the second one is that the iterative optimization involving the generator is time-consuming.

\begin{figure*}[t]
\begin{center}
\begin{tabular}{c}
    \includegraphics[scale=0.835]{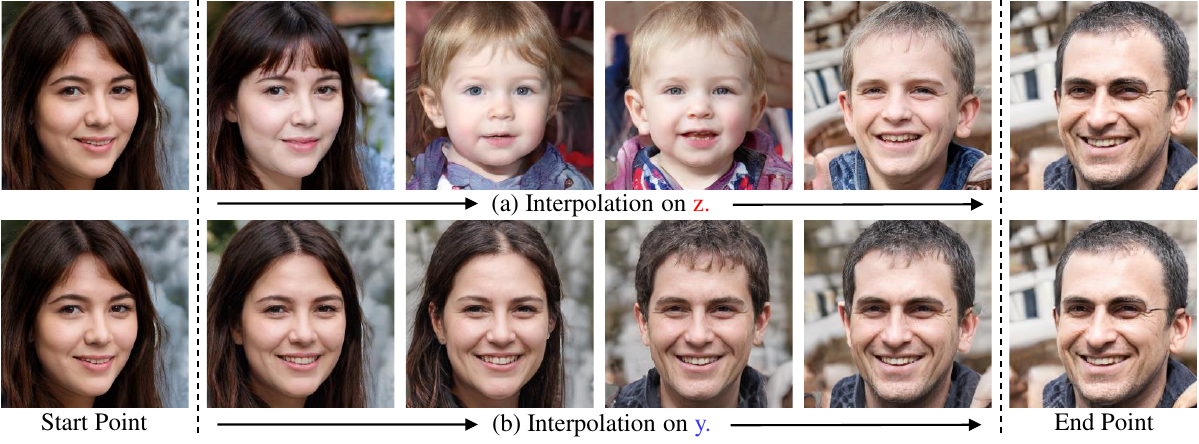}\\
    \includegraphics[scale=0.5]{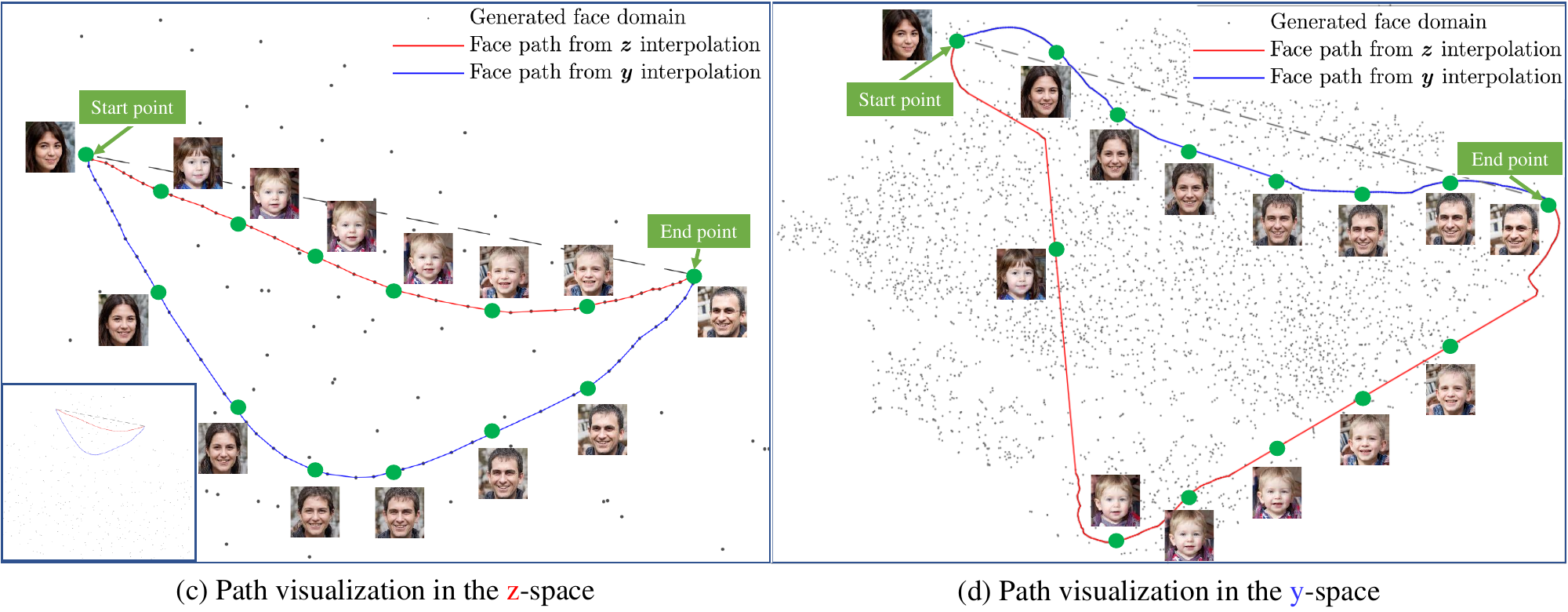}\\
\end{tabular}
\end{center} \vspace{-0.5cm}
  \caption{Illustration of the disentanglement in latent spaces of GANs. (a) and (b) show the interpolation results in the $\redz$ and $\bluey$ spaces between two images, respectively . For (c) and (d), we randomly sample 4,000 faces (including the two faces used in (a) and (b)) as the face domain (gray dots) that is embedded using t-SNE~\citep{tSNE2008}, and then find the two images' interpolated paths (red/blue curves) in the $\bm z$-space and $\bm y$-space respectively.
  The left-bottom window in (c) shows the zoom-out view of the entire $\bm z$ domain since the whole domain is very sparse.}
  \vspace{-0.4cm}
\label{fig:entanglement}
\end{figure*}

\section{Disentanglement in GANs}\label{sec:entanglement}
We first discuss the disentanglement in GANs and its implication for image reconstruction, as the foundation of the proposed algorithm.

\subsection{Motivation}
Our motivation of learning an encoder for a GAN model via latently invertible autoencoder comes from the principle of manifold learning~\citep{Tenenbaum00,Roweis00} which has been widely used for nonlinear dimensionality reduction. 
A basic assumption about manifold learning is  that data $\bm x \in \mathbb{R}^{d_x}$ lies in an underlying manifold $\mathcal{M}^{d_m}$, where $\mathbb{R}^{d_x}$ is usually called the ambient space of $\mathcal{M}^{d_m}$, $d_x$ is the ambient dimension, and $d_m$ is the dimension of the manifold. 
For example,  $d_x = 1024\times 1024 \times 3$ for a RGB image of resolution $1024\times 1024$. $d_m$ is usually unknown but satisfies $d_m \ll d_x$. 
Figures~\ref{fig:toy}(c) and (f) illustrate the Swill-roll example commonly used in manifold learning. Based on manifold learning,  nonlinear dimensionality reduction can be carried out by the following mapping
\begin{equation}
     f: ~~\bm x \in \mathcal{M}^{d_m} \overset{f}{\longmapsto} \bm y \in \mathbb{R}^{d_y},
\end{equation}
where $\bm y $ is the associated $d_y$-dimensional representation of $\bm x$.
To obtain $\bm y $ with desirable properties, some geometric constraints may be imposed on $f$, such as the geodesic distance between arbitrary pairwise points~\citep{Tenenbaum00}, locally linear fitting~\citep{Roweis00}, geometric adjacency via graphs~\citep{Belkin2003}, and geometric alignment via local tangent spaces~\citep{LTSA}. Our work is inspired by manifold learning of preserving geodesic distances.

The global geometry of data will be well maintained after nonlinear dimension reduction $f(\bm x)$ if the distance of the shortest path (approximate geodesic) between $\bm x_i$ and $\bm x_j$ is equal to that of the shortest path between $\bm y_i$ and $\bm y_j$~\citep{Tenenbaum00}. This is also called isometric mapping, as shown in Figures~\ref{fig:toy}(c) and (e). This simple but elegant geometric property is found to be useful in designing recent GAN models. Particularly \citep{StyleGAN18,StyleGAN2019} find that the consistency between interpolation paths in the latent space and the image space directly influences the performance of generation. Based on this critical observation, they define the concept of disentanglement in GANs via perceptual path length (PPL)~\citep{Zhang2018PPL}.
In this work, the nonlinear mapping $f(\bm x)$ for dimensionality reduction is an encoder parameterized by a deep neural network for inference or inversion of a GAN model. The quality of the encoder depends on the geometry of the latent space and the image manifold. In GANs, the corresponding property is named disentanglement.  Here, we follow the works by~\citep{Tenenbaum00} and~\citep{StyleGAN18,StyleGAN2019} to study the problem of adding encoders to GANs from the geometric point of view, which is rarely explored by previous works. Overall, we will say that the GAN generation obtains disentanglement if it admits an isometric mapping between the latent space and the generated image space. Otherwise, it is called entanglement.

To better illustrate the meaning of disentanglement in GANs, we use a motivating example, the Swiss-roll manifold in Figure~\ref{fig:toy} that is commonly harnessed in manifold learning. A good functional mapping for manifold embedding is that a straight blue line in the coordinate space ($\bm y$-space) corresponds to the shortest blue path on the manifold, as Figures~\ref{fig:toy}(b) and (c) display. The ideal case is the isometric mapping in geometry, meaning that the distance between $\bm y_i$ and $\bm y_j$ is identical to that between $g(\bm y_i)$ and $g(\bm y_j)$. The Isomap algorithm~\citep{Tenenbaum00} in manifold learning is established on this geometric property. For generative models like GAN, this property implies that the path from $g(\bm y_i)$ to $g(\bm y_j)$ on the image manifold reflects the smooth image deformation like face interpolation in Figure~\ref{fig:entanglement}(b). This property represents the disentanglement we use in this paper, which is consistent with the meaning in~\citep{StyleGAN18,StyleGAN2019}. On the contrary, if the straight line in the coordinate space is mapped onto an arbitrary curve on the manifold such as the red curve in Figure~\ref{fig:toy}(c), then the path between $g(\bm y_i)$ and $g(\bm y_j)$ on the manifold may incur some  uncertain properties. For GAN models, the irrelevant instances may occur on the interpolation path between $g(\bm y_i)$ to $g(\bm y_j)$, as displayed by the interpolated faces in  Figure~\ref{fig:entanglement}(a). This exhibits the property of entanglement. 
For the Swiss-roll manifold, the disentangled path corresponds to the nearly straight line in the coordinate space and the entangled path is the red curve in Figure~\ref{fig:toy}(e), which illustrates the disentanglement of latent spaces that we will study in GANs.

\setlength{\tabcolsep}{14.5pt}
\begin{table}[t]
\centering
\caption{Perceptual path length in the $\bm z$-space and $\bm y$-space.} \vspace{-0.1cm}
  \begin{tabular}{l | c | c}
        \toprule
                     &  $\bm z$-space   &   $\bm y$-space       \\
        \midrule
        Full         &      40.25       &   \textbf{23.26}       \\
        \bottomrule 
  \end{tabular}\label{tab:Disentanglement}
  \vspace{-0.2cm}
\end{table}

\subsection{Disentanglement of Latent Spaces}
To visualize the disentanglement in the GAN model for face synthesis, we randomly sample $\bm z_i$ and $\bm z_j$, and then linearly interpolate them.  The corresponding generated faces can be obtained by  $\tilde{\bm x} = g(\bm z)$. As visualized in Figure~\ref{fig:entanglement}(a) via StyleGAN on the FFHQ database, the face identities are significantly changed through the interpolation between face $\tilde{\bm x}_i$ and face $\tilde{\bm x}_j$. For example, we can clearly see the face of a young boy emerges on the interpolation path between $\tilde{\bm x}_i$ and $\tilde{\bm x}_j$, indicating that the faces in the $\tilde{\bm x}$-space do not change in a smooth and linear way in the $\bm z$-space.
As a comparison, if the mapping network is first applied to yield two intermediate latent variables $ \bm y_i = \varphi(\bm z_i) $ and  $ \bm y_j = \varphi(\bm z_j) $, then the corresponding generated faces $\tilde{\bm x} = g(\bm y)$ by \textit{linearly} interpolating between $\bm y_i$ and $\bm y_j$ vary smoothly without obvious change of face identity (Figure~\ref{fig:entanglement}(b)), as opposed to the case in the $\bm z$-space.
To further reveal the underlying geometry, we illustrate the paths of generated faces in the image space via t-SNE~\citep{tSNE2008}. 
Figure~\ref{fig:entanglement}(d) shows that the face path associated with $\bm y$ interpolation approaches the straight line much closer than that with $\bm z$ interpolation. 
The instance in Figure~\ref{fig:entanglement}(c) is analogous to the Swiss-roll case shown in  Figure~\ref{fig:toy}(d) and (e).
In fact, this is a common phenomenon for $\bm z$ and $\bm y$ interpolations between two arbitrary faces with large variation, which is supported by quantitative comparison provided in Table~\ref{tab:Disentanglement}.  The metric~\citep{StyleGAN18} is the average perceptual path length defined as
\begin{equation}
    \ell_y = \mathbb{E}[ \frac{1}{\epsilon^2} \text{dist}\left( g( \text{lerp}(\bm y_i, \bm y_j; t) ), 
    g(\text{lerp}(\bm y_i, \bm y_j; t+\epsilon) ) \right)],
\end{equation}
where lerp means linear interpolation, dist is the perceptually-based pairwise image distance~\citep{Zhang2018PPL} with $t \sim U(0,1)$ and $\epsilon = 10^{-4}$. 
From Table~\ref{tab:Disentanglement}, we see  that the path length in the $\bm y$-space is much shorter than that in the $\bm z$-space, verifying the fact that the $\bm y$-space is more disentangled.
According to the interpretation presented in the preceding section, we call the folded $\bm z$-space is the latent space of entanglement and that the intermediate $\bm y$-space is of disentanglement.

The entanglement for $\bm z$ incurs from GAN training with \textit{random sampling} in the $\bm z$-space, because there is no geometric constraint to guarantee the geometric correspondence between $\bm z$ and $\tilde{\bm x}$. A consequence is that the spatial locations of the associated $\bm z_i$ and $\bm z_j$ are not necessarily adjacent if face  $\tilde{\bm x}_i$ and face $\tilde{\bm x}_j$ are perceptually similar. This mismatch will lead to the difficulty of GAN inference and inversion, which will be further analyzed in section~\ref{se:difficulty}. Actually, preserving the geometric adjacency is the main goal of the manifold learning~\citep{Tenenbaum00,Roweis00,Belkin2003,LTSA}.
We may also impose the geometric constraint on $\bm z$ to align its geometry with that of the image manifold. 
As reported in~\citep{StyleGAN18}. However,  a simple mapping network $\varphi$ is sufficient to establish a disentangled  latent space via  $\bm y = \varphi(\bm z)$ embedded in GANs, as already compared in Table~\ref{tab:Disentanglement}.  The functional role of $\varphi(\bm z)$ is to reshape $\bm z$ to approach  geometry-consistent coordinates of the corresponding image manifold~\footnote{In StyleGAN2~\citep{StyleGAN2019}, the authors employ the Jacobian regularizer on the $\bm y$-space to enforce an isometric mapping, thus acquiring a better disentangled $\bm y$-space. The motivation behind coincides with ours for GAN inference. }, as shown in Figures~\ref{fig:toy}(a) and (b). Accordingly, the disentanglement admits the geometric consistency between the latent space and the generated sample space shown in Figure~\ref{fig:entanglement}, which is more favorable to the inference task to be shown in section~\ref{se:difficulty}. Therefore, we devise our algorithm for GAN inference based on the disentangled $\bm y$-space instead of the $\bm z$-space.

\begin{figure*}[t]
\begin{center}
\begin{tabular}{cc}
\multicolumn{2}{c}{
\includegraphics[scale=0.4]{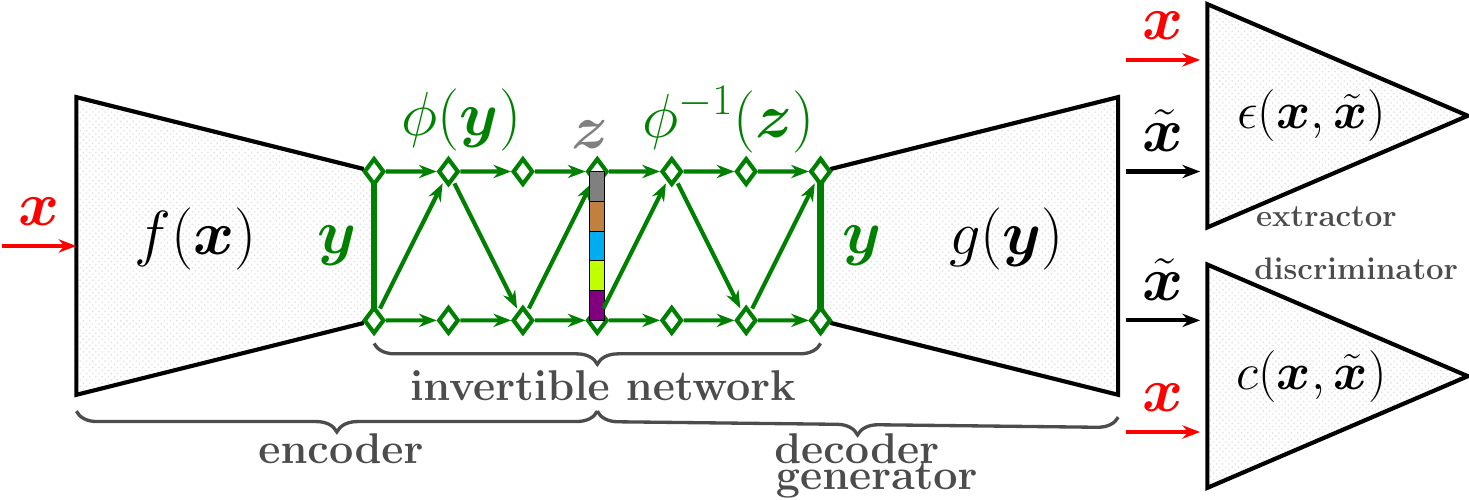}} \\
\multicolumn{2}{c}{
                              (a) Neural architecture of LIA.} \vspace{-0.3cm}                   \\ 
\includegraphics[scale=0.4]{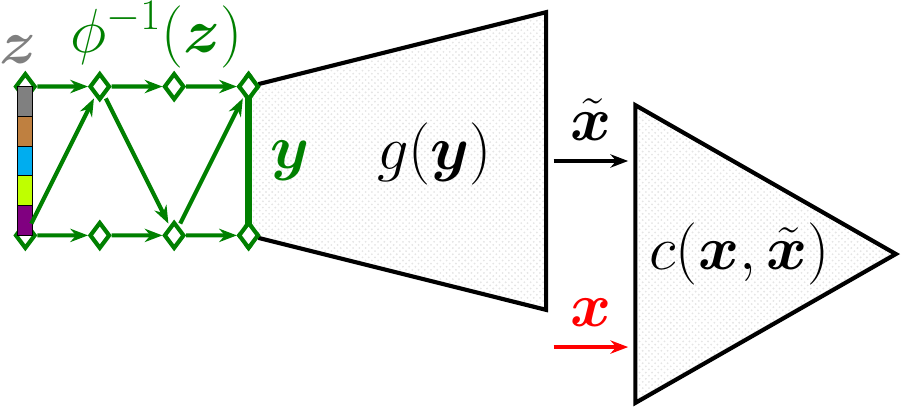}  &  \includegraphics[scale=0.4]{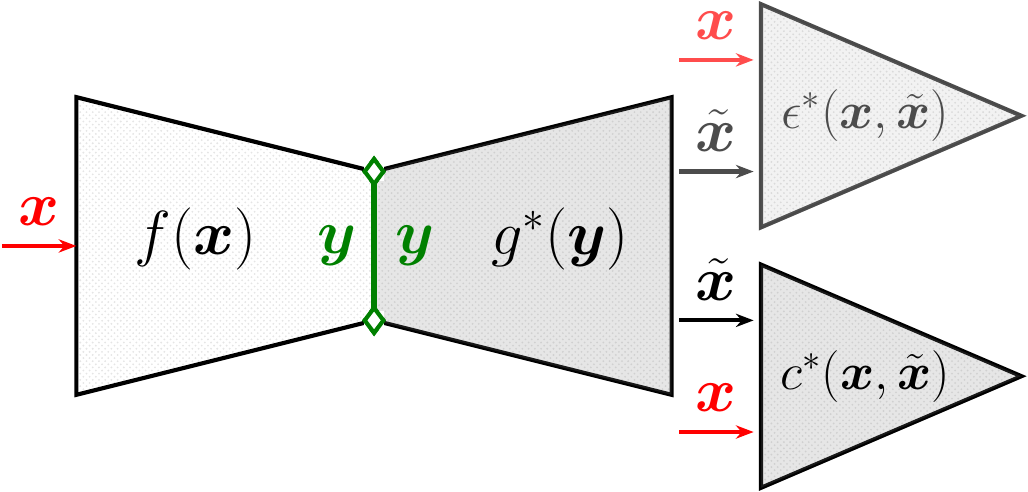} \\
             (b) Decoder training.             &          (c)  Encoder training.                 \\   

   \end{tabular}
\end{center}\vspace{-0.5cm}
   \caption{Latently invertible autoencoder (LIA) with adversarial learning. (a) LIA consists of five functional modules: an encoder to extract features $\bm y = f(\bm x)$, an invertible network $\phi$ to reshape feature embeddings to match the prior distribution $\bm z = \phi(\bm y)$ and $\phi^{-1}$ to map latent variables to disentangled feature vectors $\bm y = \phi^{-1}(\bm z)$, a decoder to produce output $\tilde{\bm x} = g(\tilde{\bm y})$, a feature extractor $\epsilon$ to perform reconstruction measure, and a discriminator $c$ to distinguish real/fake distributions. The training of LIA proceeds in the two-stage way: (b) first training the decoder via a GAN model  and (c) then the encoder by detaching the invertible network from LIA. The parameters of modules in dark gray in (c) are frozen in this stage.  } \vspace{-0.3cm}
\label{fig:LIA-net}
\end{figure*}

\subsection{Inference Problem in the $\bm z$-Space}\label{se:difficulty}
In general,  the optimization of the objective $L_g(\bm z)$ with respect to $\bm z$ can be solved with gradient decent as
\begin{equation}
    \bm z^{t+1} =  \bm z^{t} - \nabla_{\bm z} L_g(\bm z^t),
\end{equation}
where $\nabla_{\bm z} L_g(\bm z^t)$ denotes the gradient of $L_g(\bm z)$ of step $t$. Here $\bm z$ may be the prior used in GANs or the output of the encoder in VAE, $i.e.$ $\bm z = f(\bm x)$. To make the problem easily understood, we suppose that the generated sample $\tilde{\bm x}=g(\bm z)$ is also a face. 
For the entangled $\bm z$-space, the $\bm z^t$-path will probably depart for the optimal one (usually disentangled path) during the optimization process shown in Figure~\ref{fig:entanglement}(d). The emergence of irrelevant faces shown in Figure~\ref{fig:entanglement}(a) will mislead the optimization towards uncertain purpose because the objective significantly changes during the process,  thus leading to the difficulty that the algorithm may converge at the sub-optimal minima. In section~\ref{exp:dis-comparison} we will demonstrate this through extensive experiments. 
From the above analysis, we argue that the entanglement of the $\bm z$-space is the reason why the VAE/GAN and the MSE-based optimization with respect to $\bm z$ are incapable of well performing inference.

\section{Latently Invertible Autoencoder}\label{se:LIA}
As analyzed in the preceding section, to acquire the disentangled latent space, we need to embed a mapping network in the architecture of the vanilla GAN, $i.e.$ $\bm z \overset{\varphi}{\mapsto} \bm y  \overset{g}{\mapsto} \tilde{\bm x}$. At the same time, the encoder $f$ has to directly infer $\bm y$ to favor the disentanglement. Thus we need to establish a reverse mapping to obtain $\bm z$ from $\bm y$, $i.e.$ $\bm z = \phi(\bm y)$, which implies 
\begin{equation}
    \varphi(\bm z) = \phi^{-1} (\bm z).
\end{equation}
To accommodate the need, an invertible neural network can establish the reversibility between $\bm z$ and $\bm y$. Thus we design the LIA framework with the invertible mapping in the latent space. Then a two-stage training scheme is proposed to preserve the image synthesis quality and the invertibility. The details are described below.

\subsection{Neural Architecture of LIA}
 As shown in Figure~\ref{fig:LIA-net}(a), we embed an invertible neural network into the latent space in a symmetric way, following the diagram of mapping process as 
\begin{equation}\label{eq:AE-NF} 
\overbrace{\bm x \overset{f}{\longmapsto} \underbrace{\bm y \overset{\phi}{\longmapsto} }_{\text{invertible}} }^{\text{encoder}} \bm z \overbrace{\underbrace{ \overset{\phi^{-1}}{\longmapsto} {\bm y}}_{ \text{invertible}} \overset{g}{\longmapsto} \tilde{\bm x}}^{ \text{decoder}}, 
\end{equation} 
where $\phi$ denotes the deep composite mapping of the invertible network.
LIA first performs nonlinear dimensionality reduction on the input data $\bm x $ and transforms the output into the low-dimensional disentangled feature space $\mathbb{R}^{d_y}$. The role of $f(\bm x)$ in LIA can be regarded to unfold the underlying data manifold. 
Therefore, Euclidean operations such as linear interpolation and vector arithmetic are more reliable and continuous in this disentangled feature space. Then we establish an invertible mapping $\phi(\bm y)$ from the feature $\bm y$ to the latent variable $\bm z$, as opposed to VAEs that directly map original data to latent variables. The feature $\bm y$ can be exactly recovered via the invertibility of $\phi$ from $\bm z$, which is the strength of using invertible networks. The recovered feature $\bm y$ is then fed into a partial decoder $g(\bm y)$ to generate the corresponding data $\tilde{\bm x}$. If the maps  $\phi$ and  $\phi^{-1}$ of the invertible network are bypassed, LIA reduces to a standard autoencoder, $i.e.$ $\bm x \overset{f}{\mapsto} \bm y  \overset{g}{\mapsto} \tilde{\bm x}$.

In general, any invertible networks are applicable in the LIA framework. We find in practice that a simple invertible network called NICE used in~\cite{NICE15} is sufficiently capable of constructing the mapping from the feature space $\mathbb{R}^{d_y}$ to the latent space $\mathbb{R}^{d_z}$. Let $\bm x = [ \bm x_t ; \bm x_b ]$ and $\bm z = [ \bm z_t ; \bm z_b ]$ be the forms of the top and bottom fractions of $\bm x$ and $\bm z$, respectively. Then the invertible network can be built as 
\begin{align}
    \bm z_t &= \bm x_t, \quad \bm z_b = \bm x_b + \tau(\bm x_t),\\
    \bm x_t &= \bm z_t, \quad \bm x_b = \bm z_b - \tau(\bm z_t), 
\end{align}
where $\tau$ is the transformation that can be an arbitrary differentiable function. Alternatively, one can attempt to exploit the complex invertible network with affine coupling mappings for more challenging tasks~\citep{NVP17,Glow18}. As conducted in~\cite{NICE15}, we set $\tau$ for simplicty as a multi-layer perceptron with the leaky ReLU activation.
\subsection{Reconstruction Loss and Adversarial Learning}
To guarantee the precise reconstruction $\tilde{\bm x}$, the conventional way by (variational) autoencoders is to use the distance $\| \bm x- \tilde{\bm x}\|$ or the cross entropy directly between $\bm x$ and $\tilde{\bm x}$. Here, we utilize both the pixel loss and the perceptual loss that is proven to be more robust to variations of image details~\citep{perceptualLoss2016}. Let $\epsilon$ denote a feature extractor. Then we can write the loss 
\begin{equation}\label{eq:lia-rec-loss}
    L(\epsilon,\bm x, \tilde{\bm x}) = \| \bm x - \tilde{\bm x} \| + \beta_1 \| \epsilon(\bm x) - \epsilon(\tilde{\bm x}) \|. 
\end{equation}
where $\beta_1$ is the hyper-parameter to balance those two losses. The feasibility for this type of mixed reconstruction loss is actually evident in diverse image-to-image translation tasks. 
It suffices to emphasize that the functionality of $\epsilon$ here is in essence to produce the representations of the input $\bm x$ and the output $\tilde{\bm x}$. The acquisition of $\epsilon$ is fairly flexible.  It can be attained by supervised \textit{or} unsupervised learning, meaning that $\epsilon$ can be trained with class labels like VGG~\citep{VGG} or without class labels~\citep{CPC2018}.  

The norm-based reconstruction constraints usually incur the blurry output images in the autoencoder-like architectures~\citep{noise2noise}.  This problem can be handled via adversarial learning~\citep{GAN}. To do so, a discriminator $c$ is employed to balance the loss of the distribution comparison between $\bm x$ and $\tilde{\bm x}$. Here we can use the original non-saturating loss $V(g,c)$ in Equation~(\ref{eq:gan-loss}) or the loss of Wasserstein distance~\citep{WGAN17,WGAN-GP17}, $i.e.$
\begin{equation} \label{eq:lia-wgan-loss}
  L(c)=  \mathbb{E}_{p(\tilde{\bm x})}[c(\tilde{\bm x})] - \mathbb{E}_{p(\bm x)}[c(\bm x)] +~\gamma\mathbb{E}_{p(\bm x)}\big[  \|\nabla_{{\bm x}}c({\bm x})\|^2 \big], %
\end{equation}
where $\gamma$ is the hyper-parameter of the gradient regularizer~\citep{whereGAN18}. In practice, the sliced Wasserstein distance that is approximated by Monte Carlo sampling is preferred to compute the distance between  $p_x$ and $p_{\tilde{x}}$~\citep{PGGAN}.

\section{Two-Stage Training}\label{se:training}
 We propose a two-stage training scheme, which decomposes the framework into two parts that can be well trained end-to-end respectively, as shown in Figure~\ref{fig:LIA-net}(b) and (c). First, the decoder of LIA is trained as a GAN model with the invertible network. Second, the invertible network that connects the feature space and the latent space is detached from the architecture, reducing the framework to a standard autoencoder \textit{without} variational inference.  Thus this two-stage scheme makes the reconstruction happen in the disentangled latent space while avoiding the issue commonly encountered in VAE~\citep{Lucas2019Collapse}.
\subsection{Decoder Training}
ProGAN~\citep{PGGAN}, StyleGAN~\citep{StyleGAN18,StyleGAN2019}, and BigGAN~\citep{BigGAN} are capable of generating photo-realistic images from random noise sampled from some prior distributions. Then it is naturally supposed that such GAN models are applicable to recover a precise $\tilde{\bm x}$ if we can find the latent variable $\bm z$ for the given $\bm x$. Namely, we may train the associated GAN model separately in the LIA framework. To conduct this, we single out a standard GAN model for the first-stage training, as displayed in Figure~\ref{fig:LIA-net}(b), 
the diagram of which can be formalized by 
\begin{equation}\label{eq:diagramGAN}
\bm z \overset{\phi^{-1}}{\longmapsto}  \bm y \overset{g}{\longmapsto} \tilde{\bm x} \leftrightarrow 
\Big \{
\begin{matrix}
  \tilde{\bm x} \\
  \bm x
 \end{matrix} \Big\}
\overset{c}{\longmapsto}  V(g,c), 
\end{equation}
where $\bm z$ is directly sampled from a pre-defined prior. According to the principle of the original GAN, the optimization objective can be written as 
\begin{equation}
    \{\phi^{*},g^{*},c^{*}\} = \min_{\phi,g}\max_{c}  V(g,c).
\end{equation}
It is worth noting that the role of the invertible network here is just its transformation invertibility. \textit{We do not pose any constraints on the probabilities of $\bm z$ and $\phi(\bm y)$ in contrast to normalizing flows}.
Our strategy of attaching an invertible network in front of the generator can be potentially applied to any GAN models. 

\subsection{Encoder Training}
In the LIA architecture, the invertible network is embedded in the latent space in a symmetric fashion, $i.e.$ $f(\bm x)= \bm y = \phi^{-1}(\bm z)$. This unique characteristic of the invertible network allows us to detach the invertible network $\phi$ from the LIA framework. Thus we attain a conventional autoencoder without stochastic variables, as shown in Figure~\ref{fig:LIA-net}(c). 
 We can write the diagram
\begin{equation}\label{eq:diagramAEA}
\bm x \overset{f}{\longmapsto}  \bm y \overset{g^{*}}{\longmapsto} \tilde{\bm x} \leftrightarrow
\Big \{
\begin{matrix}
  \tilde{\bm x} \\
  \bm x
 \end{matrix} \Big\}
\begin{matrix} 
   \overset{\epsilon^{*}}{\longmapsto} L(\epsilon^{*},\bm x, \tilde{\bm x}) \vspace{-0.0cm}\\
   \overset{c^{*}}{\longmapsto}V(g^{*},c^{*})\text{ }\text{ } 
\end{matrix}.
\end{equation}
In practice, the feature extractor $\epsilon$ in the perceptual loss is the VGG features pretrained on the ImageNet dataset. After the first-stage GAN training, the parameter of $f$ is learned as 
\begin{equation}\label{eq:encoder-loss}
    \{f^{*},\hat{c}^{*}\} = \min_{f}\max_{c^{*}} V(g^{*},c^{*}) + \beta_2 L(\epsilon^{*},\bm x, \tilde{\bm x}), 
\end{equation}
where $\beta_2$ is the hyper-parameter and $\hat{c}^{*}$ is the fine-tuned parameter of the discriminator, meaning that the discriminator is fine-tuned with the training of the encoder while the generator is frozen. The above optimization serving to the architecture in Figure~\ref{fig:LIA-net}(c) is widely applied in computer vision. It is the backbone framework of various GANs for diverse image processing tasks~\citep{pix2pix,cycleGAN}. For LIA, however, it is much simpler because we only need to learn the encoder $f$. The the invertible network and the two-stage training enforces the encoder to converge with more precise inference. The results will be shown in section~\ref{sec:experiment}.
\begin{figure*}[t]
\begin{center}
\begin{tabular}{c}
    \includegraphics[scale=0.82]{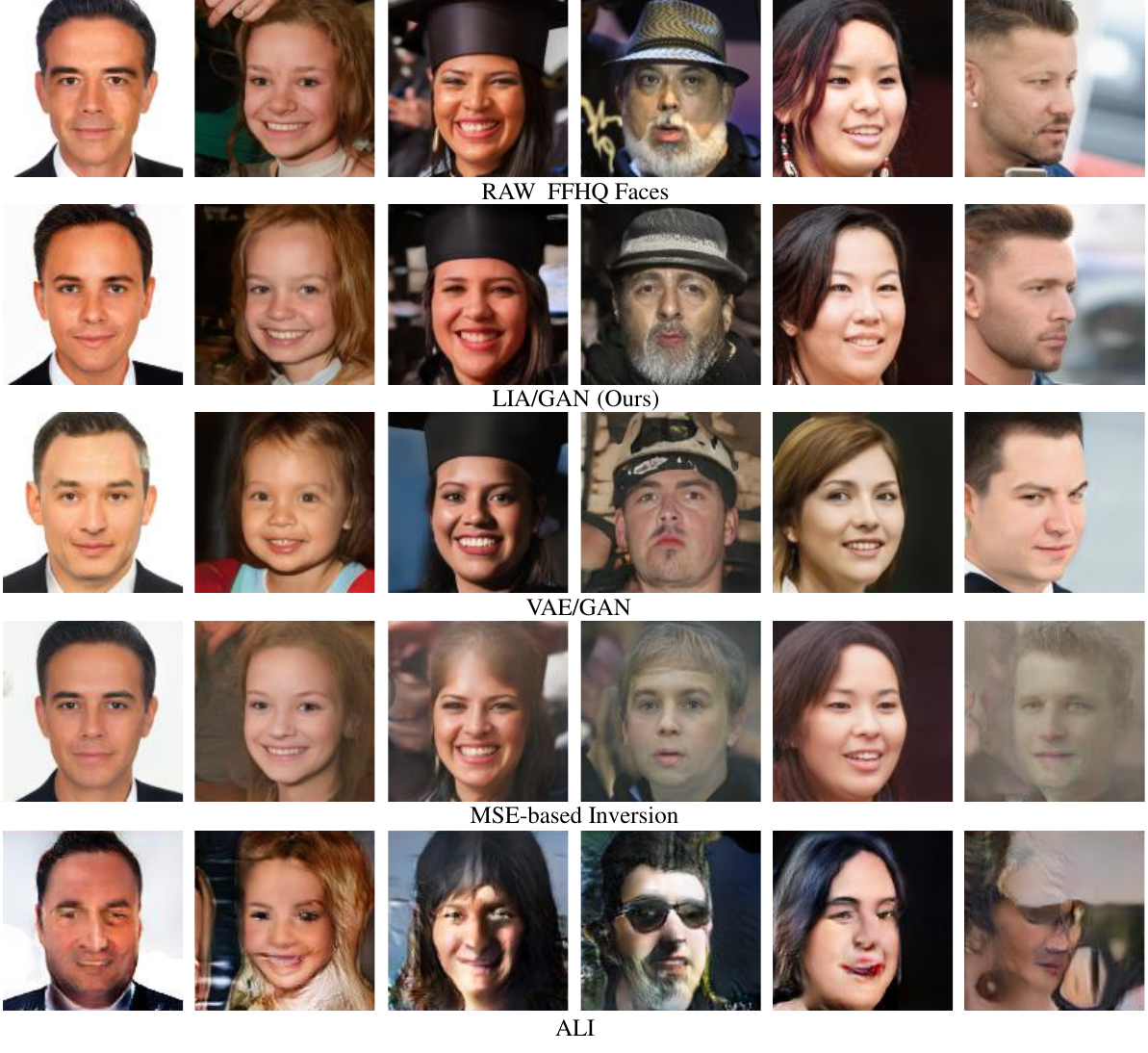}
   \end{tabular}
\end{center} \vspace{-0.7cm}
   \caption{Comparison of different methods on face reconstruction. The first row is original images in the FFHQ dataset and the rest rows are the different methods we compare. We can see that LIA produces better results in terms of image quality and reconstruction accuracy, where the age, hat, and pose, are all well preserved. ALI is not suitable in this scenario because it conveys high-level semantic information which is more powerful for recognition.}
   \vspace{-0.4cm}
\label{fig:reconstruction}
\end{figure*}
\section{Related Work}\label{sec:related-work}
Our LIA model is relevant to the works that solve the inference problem for VAEs with adversarial learning as well as the works that design encoders for GANs. 
The integration of GAN with VAE can be traced back to the work of VAE/GAN~\citep{VAEGAN} and the implicit autoencoders~\citep{Makhani15,Makhani18}. 
These methods encounter the difficulty of end-to-end training, because the gradients are prone to become unstable after going through latent spaces in deep complex architectures~\citep{Bowman15,Kingma16}. 
Besides, there is an intriguing attempt of training VAE in the adversarial manner~\citep{Ulyanov17,pioneerNet18}. 
These approaches confront the trade-off between the roles of the encoder that performs inference and compares the real/fake distributions. 
This is difficult to tune. So we prefer the framework of the vanilla GAN with an indispensable discriminator.

The works relevant to LIA are the models of combining VAE and the inverse autoregressive flow~\citep{Kingma16} and the latent-flow-based VAE approach that are VAEs with latent variables conditioned by normalizing flows~\citep{fVAE18,GLF2019}. 
These three models all need to optimize the log-likelihood of normalizing flows, which is essentially different from  LIA. 
The invertible network in LIA only serves to establish the invertibility between the $\bm z$-space and the disentangled $\bm y$-space. 
There is no probabilistic optimization for normalizing flows involved in LIA.
There are alternative attempts of specifying the generator of GAN with normalizing flow~\citep{flowGAN17} or mapping images into feature spaces with partially invertible network~\citep{Lucas2019}. 
These approaches suffer from high complexity computation for high dimensions. 
The approach of two-stage training in~\cite{inverseGenerator17} suffers the entanglement problem.

It is worth noting that the reconstruction task we focus here is different from the recent work of representation learning which learns features for recognition and classification using adversarial inference~\citep{ALI2017,BiGAN2017, BigBiGAN2019}. 
Our primary goal is to infer latent code of a real image and faithfully reconstruct it for the downstream image editing tasks. 
The performance of image editing completely depends on the reconstruction precision whereas the works of adversarial inference such as ~\cite{ALI2017,BiGAN2017, BigBiGAN2019} focus on learning high-level semantic features for classification task. 
The goals are substantially different.

\setlength{\tabcolsep}{8.5pt}
\begin{table}[t]
\centering
\caption{Quantitative comparison of image reconstruction on the test dataset.  $\downarrow$ means that the lower numbers are better.}\label{tab:accuracyConstruct} 
\vspace{-0.2cm}
  \begin{tabular}{  l | c  c  c  c }
    \toprule
    Metric    & LIA/GAN & ALI & MSE & VAE/GAN\\
    \midrule
    \FID & \textbf{19.26}  & 74.98  & 44.79  & 22.26 \\
    \midrule
    \SWD & \textbf{12.16}  & 15.09  & 43.44  & 15.82 \\
    \midrule
    \MSE & \textbf{11.79}  & 32.61  & 18.81  & 23.18 \\
    \bottomrule
  \end{tabular} \vspace{-0.5cm}
\end{table}

\begin{figure*}[t]
\begin{center}
\begin{tabular}{c}
    \includegraphics[scale=0.84]{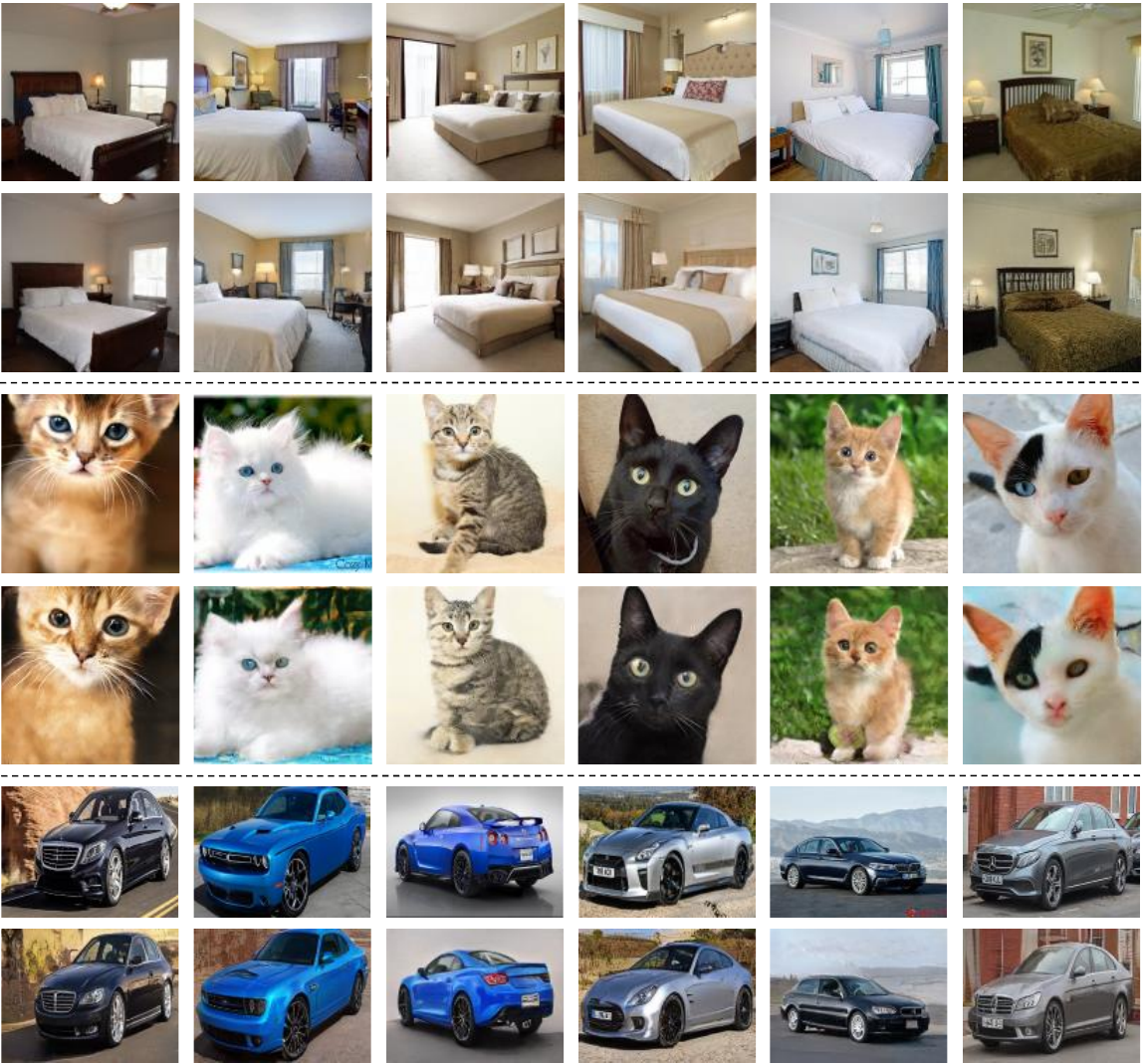}
   \end{tabular} \vspace{-0.6cm}
\end{center}
   \caption{The exemplar real images of objects and scenes from the LSUN validation set and their reconstructed images by LIA. Three categories are tested, \textit{i.e.} bedroom, cat, and car.}
   \vspace{-0.4cm}
\label{fig:reconstruction-lsun}
\end{figure*}

We are aware that a concurrent work, called Adversarial Latent Auto-Encoders (ALAE)~\citep{ALAE2020} proposed a similar idea to ours. 
There are four critical differences between ALAE and our LIA method. 
First, ALAE uses the style-based encoder that is more complex than ours. 
For LIA, there is no special constraint to the architecture of the encoder. 
Second, the encoder of ALAE is also the main module for feature extraction used in reconstruction loss and the discriminator, which allows the end-to-end training of the whole algorithm. 
However, LIA can be integrated with other GAN frameworks due to its much more flexible modular design and the scheme of the two-stage training. 
Third, we explicitly interpret and reveal the underlying reason why GAN inference needs to be performed in the deterministic  $\bm y$-space instead of the stochastic $\bm z$-space. 
Finally, the $\bm y$-space and the $\bm z$-space are exactly invertible for LIA, which provides a convenient way of investigating one from another. 
We will demonstrate the application of GAN inversion in the experiment section.

\section{Experiments}\label{sec:experiment}

\noindent\textbf{Implementation Details.}
For the experimental setup, we instantiate the decoder of LIA with the generator of StyleGAN~\citep{StyleGAN18}. 
The difference is that we replace the mapping network (MLP) in StyleGAN with the invertible network, and the layer number of the invertible network is 8.
The hyper-parameters are set as $ \beta_1=5\text{e-}5 $, $ \beta_2=0.1 $, and $ \gamma=5 $ in Equations~(\ref{eq:lia-rec-loss}), (\ref{eq:encoder-loss}), and (\ref{eq:lia-wgan-loss}), respectively. 
For perceptual loss in Equation~(\ref{eq:lia-rec-loss}), we take $\epsilon=\textsf{conv}4\_3$ from the VGG weight.

\begin{figure*}[t]
\begin{center}
    \includegraphics[scale=0.84]{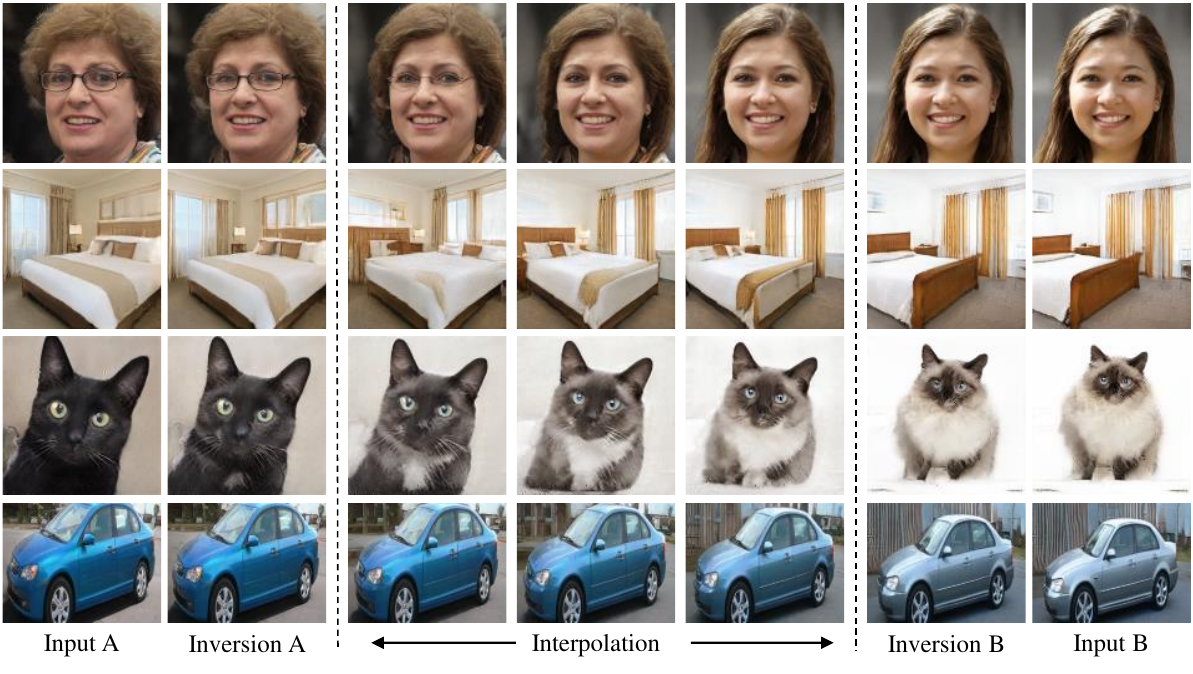}
\end{center} \vspace{-0.7cm}
   \caption{Interpolation on real images using LIA on different datasets. We can see that LIA is capable of generating smooth change for the pose, layout, and texture.}
   \vspace{-0.4cm}
\label{fig:interpolation}
\end{figure*}

\begin{figure*}[t]
\begin{center}
    \includegraphics[scale=0.73]{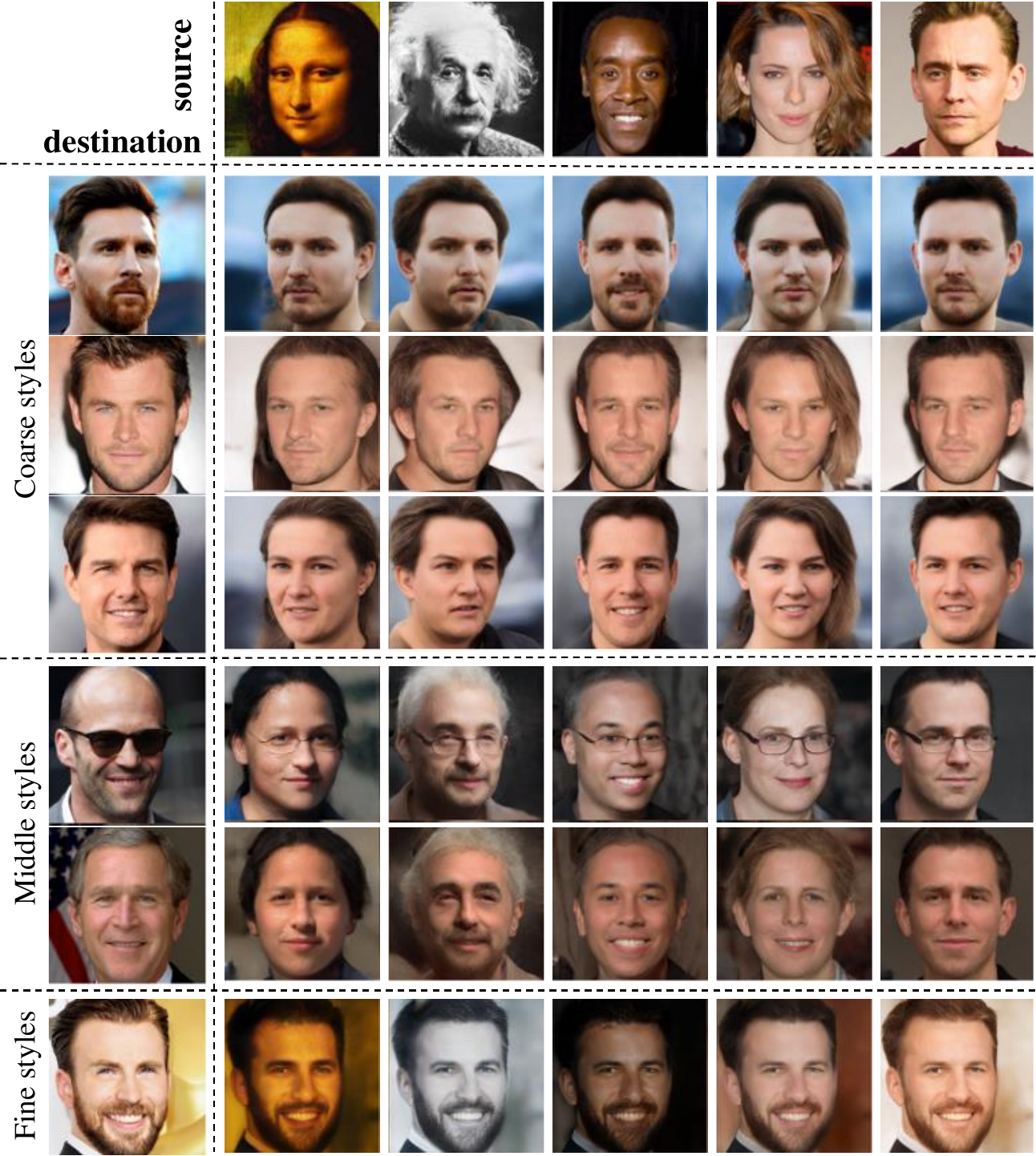}
\end{center} \vspace{-0.4cm}
   \caption{Style mixing for real faces. First column indicates the destination images and the first row shows the source images. The coarse style, the middle style, and the fine style show the mixed results with the latent codes of destination faces replaced using  the latent codes of source faces at resolution $4^2$-$8^2$, $16^2$-$32^2$, $64^2$-$128^2$, respectively. }
\vspace{-0.3cm}
\label{fig:stylemixing}
\end{figure*}

\begin{figure*}[ht]
\begin{center}
    \includegraphics[scale=0.85]{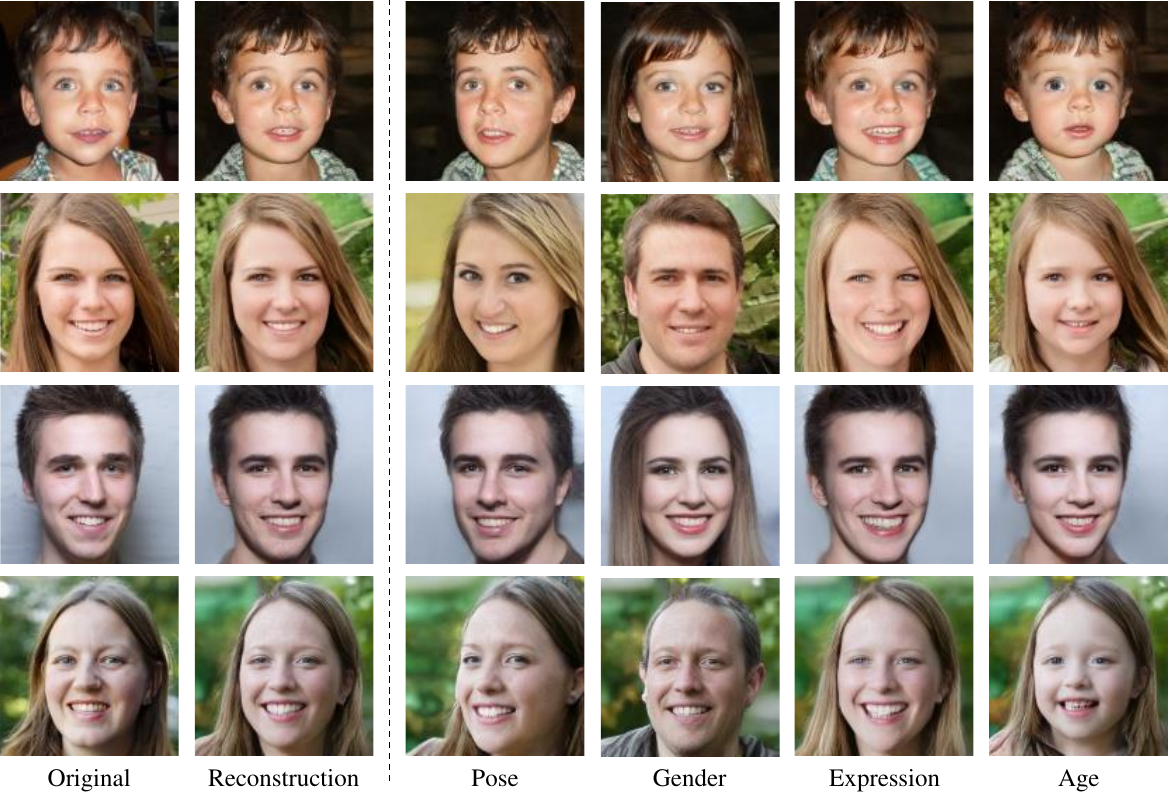}
\end{center} \vspace{-0.6cm}
   \caption{Manipulating real faces through varying the inverted latent codes. The first column is the input images, the second column is the inversion results obtained using LIA, and the rest columns are the editing results.}
   \vspace{-0.3cm}
\label{fig:manipulation}
\end{figure*}

\noindent\textbf{Datasets and Metrics.}
We evaluate our method on four datasets.
The first one is Flickr-Faces-HQ (FFHQ) Database~\citep{StyleGAN18}, and the remaining three come from the LSUN database~\citep{LSUN15}, $i.e.$, bedroom, cat, and car.
For FFHQ, we take the first 65,000 faces as the training set and the remaining 5,000 faces as the reconstruction test according to the exact order of the dataset.
We do not split the dataset by random sampling for interested readers can precisely reproduce all the reported results with our experimental protocol.
And for the datasets in LSUN, the 0.1 million images are selected by ranking algorithm~\citep{Zhou:2003} from the first 0.5 million images in the dataset. 
Each cat and bedroom image is resized to be $128\times 128$, and the size of the car image is $128\times 96$ for training. 
We take subsets because it does not take too long for training to converge while still maintaining the data complexity.
For quantitative evaluation metrics, we use Fr\'{e}chet Inception Distance (FID), Sliced Wasserstein Distance (SWD), Mean Squared Error (MSE),  Structural Similarity Index Measure (SSIM), and Peak Signal-to-Noise Ratio (PSNR). 
These metrics are commonly used to measure the quality of the images in GANs~\citep{Ulyanov17, PGGAN, BiGAN2017, StyleGAN18}.

We organize the experiments as follows:
First, we validate the effectiveness of our proposed LIA on a wide range of datasets in section~\ref{exp:inference-quality-app}, and some applications are given as well.
Second, we demonstrate that the reconstruction can be further improved through finetuning the inferred code from the encoder on a specific image in section~\ref{exp:finetuning}.
Based on finetuning, we reveal that the disentanglement is the key factor of attaining high-quality GAN inversion in section~\ref{exp:dis-comparison}.

\subsection{Inference Quality and Its Applications} \label{exp:inference-quality-app}
The main goal of this paper is to endow GANs with the ability of inference, $i.e.$, projecting real samples into the latent space of GANs with faithful reconstructions.
Such ability allows us to fulfill some downstream tasks such as image editing.
In this section, we first show the inference ability of LIA on a wide range of datasets. 
Then, we demonstrate that the latent codes obtained from LIA can be used in various downstream tasks such as interpolation, style-mixing, and manipulation.

\subsubsection{Inference Quality} \label{exp:inference-quality}
Inference quality can be measured by reconstruction precision, which is an important evaluation on whether the inverted code can well represent the input image since our downstream tasks require high reconstruction precision.

\noindent\textbf{FFHQ Database.}
We first conduct experiments on FFHQ and show the comparison results with baselines in Table~\ref{tab:inference-methods}.
For MSE-based GAN inversion~\citep{DCGAN16, BEGAN17, Lipton17}, we use the code released by Puzer\footnote{\url{https://github.com/Puzer/stylegan-encoder}}. 
For adversarial inference, we compare the method named adversarially learned inference (ALI)~\citep{ALI2017}.
To evaluate the necessity of the invertible network, we also train an encoder and StyleGAN with its original multi-layer perceptron, which is the VAE/GAN approach in Table~\ref{tab:inference-methods}.
It is worth noting that the two-stage training scheme is also used as LIA does.

Figure~\ref{fig:reconstruction} shows the reconstructed faces of all the methods.
It is clear that LIA outperforms others from visual perception. The reconstructed faces by ALI look semantically plausible, but the reconstruction quality is mediocre.
The method of the MSE-based optimization produces facial parts of comparable quality with LIA when the faces are normal. 
But this approach fails when the variations of faces become large.
For instance, the failure comes from the long fair, hats, beards, and large poses.
Interestingly,  StyleGAN with variational inference (VAE/GAN) fails in recovering the target faces using the same training strategy as LIA, but it is still capable of generating nearly photo-realistic faces due to the StyleGAN generator. 
This indicates that the invertible network plays a crucial role in making LIA work. 
The quantitative result in Table~\ref{tab:accuracyConstruct} also shows the best performance of LIA.

\noindent\textbf{LSUN Database.}
We further conduct experiments on the LSUN dataset to demonstrate the generalization of LIA regarding the inference ability.
Figure~\ref{fig:reconstruction-lsun} shows that the reconstructed objects by LIA faithfully maintain the semantics as well as the appearance of the original ones. 
For example, the lights in the bedroom are preserved, and the cats' whiskers are recovered, indicating that LIA is able to recover very detailed information. 
The experimental results on FFHQ and LSUN databases verify that the symmetric design of the invertible network and the two-stage training successfully handles the issue of GAN inference, which facilitates us to complete various downstream tasks.

\subsubsection{Image Editing Applications} \label{exp:applications}
In this section, we apply the latent codes inferred from LIA to several image editing applications.

\noindent\textbf{Interpolation.}
We perform the linear interpolation between two arbitrary inverted latent codes, and then feed these interpolated codes to the generator to produce a sequence of intermediate frames between these two inputs. 
The quality of interpolated images reflects that of inferred codes.
We conduct interpolation on four datasets as shown in Figure~\ref{fig:interpolation}.
We can see that the object instances vary smoothly and stay photo-realistic.

\noindent\textbf{Style Mixing.}
We perform the style mixing on StyleGAN-based architectures.
We swap the corresponding style codes at some particular layers of the generator between two images.
Compared to the previous StyleGAN work~\citep{StyleGAN2019} that conducts style mixing on the synthesized data, we can perform it directly on the real images thanks to the inference ability of LIA.
We first attain the inferred latent codes for the source images and destination images, respectively, and then swap the corresponding codes at the pre-specified layers.
Figure~\ref{fig:stylemixing} shows that our method can successfully mix the styles of the real samples at different levels.

\noindent\textbf{Manipulation.}
The latent space of the generator has been shown to encode rich semantics~\citep{shen2020interpreting,shen2020interfacegan,yang2021semantic}, allowing semantic image manipulation by linear transformation on its latent codes.
Following InterFaceGAN~\citep{shen2020interpreting,shen2020interfacegan}, we find some meaningful directions in the latent space and then use them to manipulate the real face reconstructed from the latent code $\bm y$ derived from our LIA algorithm.
Figure~\ref{fig:manipulation} shows the facial manipulation results, in which we can see that LIA can obtain satisfactory results when manipulating the pose, gender, expression, and age.

\begin{figure*}[t]
\begin{center}
\begin{tabular}{c}
    \includegraphics[scale=0.82]{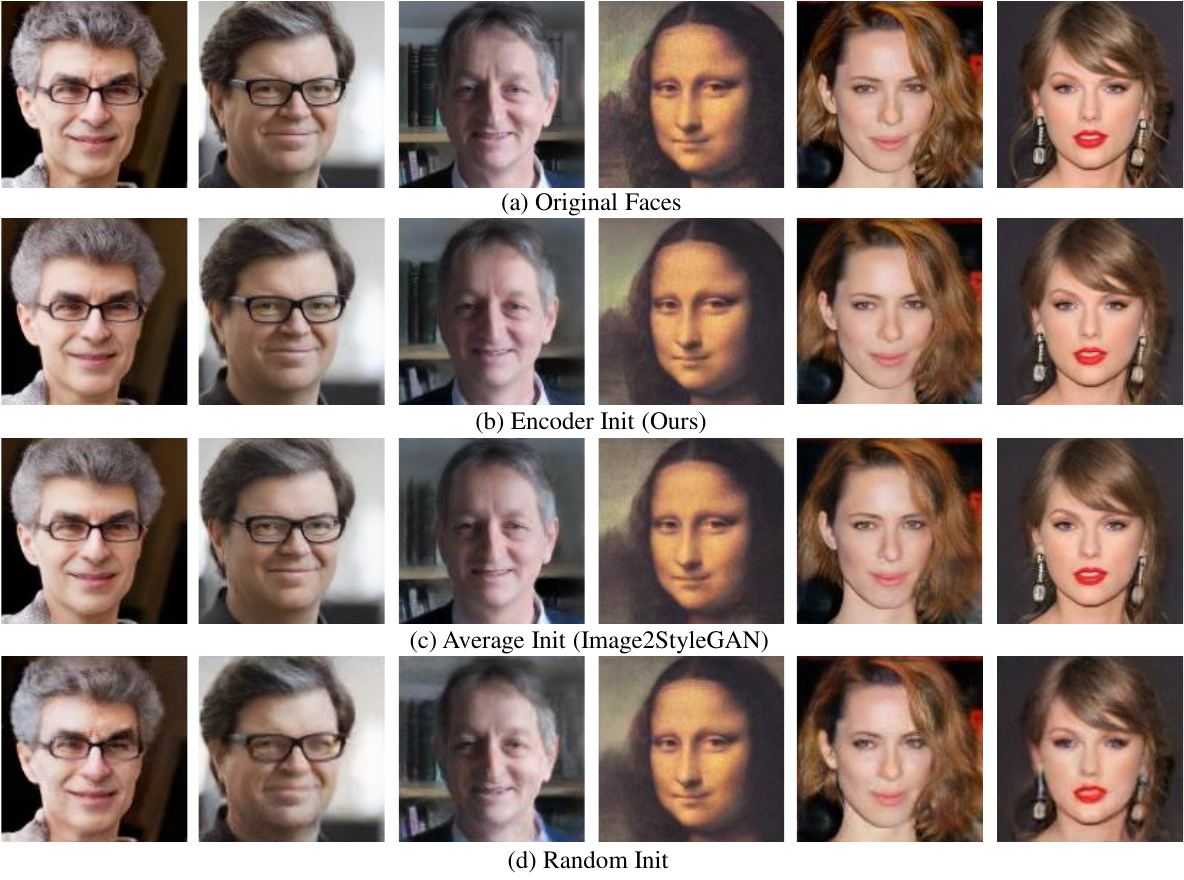}
   \end{tabular}
\end{center} \vspace{-0.7cm}
   \caption{Reconstruction from finetuning the latent code to a specific image. Three different initializations are compared. Details are shown in Figure \ref{fig:fine-tuning-detail}.}
   \vspace{-0.2cm}
\label{fig:finetuning}
\end{figure*}
\begin{figure*}[t]
\begin{center}
\begin{tabular}{c}
    \includegraphics[scale=0.82]{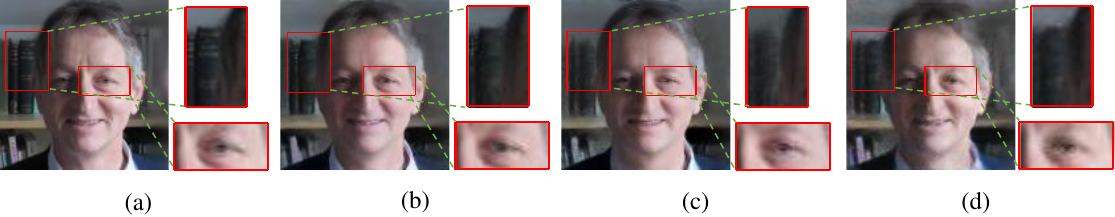}
   \end{tabular}
\end{center} \vspace{-0.6cm}
   \caption{Detailed comparison on reconstructed faces from finetuning. (a) Original faces. (b) Initialization from the output of LIA's encoder. (c) Initialization from the average (Image2StyleGAN). (d) Initialization from random sampling. We can see that the outline of the book in the background is well preserved by LIA. Rather, the outline of the rest two methods are relatively ambiguous. So is the eye.}
   \vspace{-0.4cm}
\label{fig:fine-tuning-detail}
\end{figure*}

\setlength{\tabcolsep}{8.5pt}
\begin{table}[t]
\centering
\caption{Quantitative comparison of image reconstruction using different initialization methods in the high-dimensional space. $\downarrow$ means that lower numbers are better; $\uparrow$ means that higher numbers are better.}
\vspace{-0.1cm}
  \begin{tabular}{  l | c  c  c  c}
    \toprule
       Methods      &     \MSE       &   \SSIM       &    \PSNR        &    \FID        \\
    \midrule
    Encoder Init    & \textbf{0.019} &\textbf{0.724} &\textbf{23.51}   &\textbf{39.52} \\
    \midrule
    Average Init    &   0.028        &   0.600       &    19.33        &   57.09       \\
    \midrule
    Random  Init    &   0.032        &   0.64        &    21.34        &   105.52      \\
    \bottomrule
  \end{tabular}\label{tab:quat-finetuing-high-dimension} 
  \vspace{-0.4cm}
\end{table}

\begin{figure*}[t]
\begin{center}
    \includegraphics[scale=0.82]{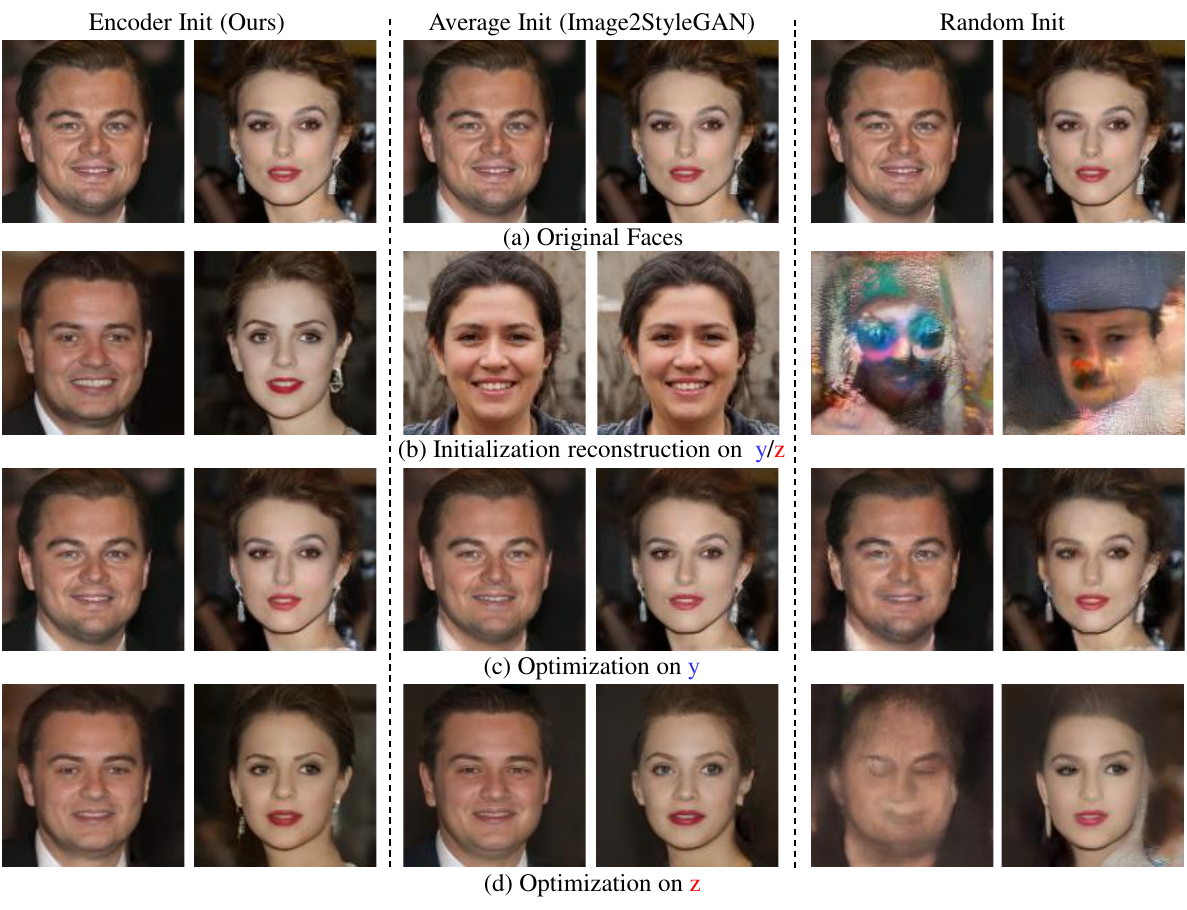}
\end{center} \vspace{-0.6cm}
   \caption{Optimization on the latent code $\bm z$ and the feature $\bm y$. The faces in the second row are reconstructed by the different initial values of $\bm z$/$\bm y$. The third row shows the result on $\bm y$ and the fourth row shows the result on $\bm z$.}
   \vspace{-0.3cm}
\label{fig:optimization_on_yz}
\end{figure*}

\setlength{\tabcolsep}{7.5pt}
\begin{table}[t]
\centering
\caption{Quantitative comparison of image reconstruction in different spaces. We use one hundred images collected from the Web as the test dataset. $\bluey$ stands for optimization in the $\bm y$-space and $\redz$  in the $\bm z$-space. Output of encoder, average value, and randomly sampled one are three different initialization methods. $\downarrow$ means that lower numbers are better; $\uparrow$ means that higher numbers are better.}
\vspace{-0.1cm}
  \begin{tabular}{  l | c  c  c  c  c}
    \toprule
         Methods           &    S     &     \MSE        &   \SSIM      &   \PSNR         &    \FID          \\
     \midrule
 \multirow{2}{*}{Encoder}  & $\bluey$ & \textbf{0.0055} &\textbf{0.86} &\textbf{29.06}   & \textbf{41.46}   \\
 \cline{2-6}
                           & $\redz$  &  0.0319         &  0.679       &    21.32        &   56.89          \\
 \midrule
 \multirow{2}{*}{Averge}   & $\bluey$ & \textbf{0.0075} &\textbf{0.83} &\textbf{27.76}   & \textbf{52.01}   \\
 \cline{2-6}
                           & $\redz$  &   0.058         &   0.620      &    18.94        &   65.86          \\
 \midrule
 \multirow{2}{*}{Random}   & $\bluey$ & \textbf{0.0081} &\textbf{0.81} &\textbf{27.18}   &  \textbf{75.34}  \\
 \cline{2-6}
                           & $\redz$  &   0.073         &   0.537      &    17.71        &   216.8          \\
 \bottomrule
  \end{tabular}\label{tab:quat-different-space} 
  \vspace{-0.4cm}
\end{table}

\subsection{Finetuning the Latent Code to a Specific Image} \label{exp:finetuning}
In the previous section, we demonstrate that the encoder of LIA is capable of achieving  satisfactory reconstruction precision.
However, the reconstruction precision can be further improved in two ways: increasing the dimension of the intermediate latent space and finetuning the latent code to fit a specific image. These two ways have been explored by Image2StyleGAN~\citep{image2stylegan}. 
This is reasonable since a larger latent space can preserve more semantic information of images, and the image-wise finetuning can recover specific features of objects or scenes.
Increasing the dimension means assigning different $\bm y$s to different layers of the StyleGAN generator rather than using the same $\bm y$s as the original StyleGAN does.
The different $\bm y$s span a high-dimensional latent space that can better unfold the underlying image manifold.
Finetuning the latent code means using Equation~(\ref{eq:gan-inversion2}) to fit the given  individual target image.
For the high-dimensional optimization problem in Equation~(\ref{eq:gan-inversion2}), a better initialization value results in a better optimization output.
The early works~\citep{Creswell16, Lipton17} use  random vectors usually sampled from a prior distribution to initialize $\bm z$ in Equation~(\ref{eq:gan-inversion2}). This kind of initialization usually leads to sub-optimal results.
 Image2StyleGAN~\citep{image2stylegan} uses the average $\bm y$ as the initial value, which significantly improves  optimization results. 
Here, we show that the initialization from the encoder of LIA can achieve better results than the aforementioned two initialization methods.
In our case, the initial value of $\bm y$ comes from the encoder of LIA when given a target image $ \bm x $, $i.e.$ $\bm y_0 = f(\bm x)$.

\begin{figure}[t]
\begin{center}
    \includegraphics[scale=0.83]{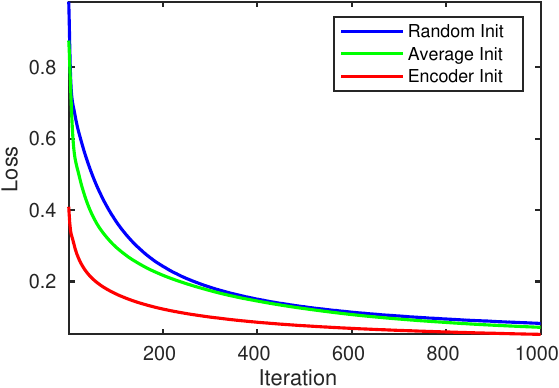}\\
\end{center} \vspace{-0.5cm}
   \caption{Reconstruction loss when finetuning different latent codes as initialization. Three initialization methods are compared: randomly sampled $\bm y$, the average of $\bm y$s (used in the original Image2StyleGAN~\citep{image2stylegan}), and the output code $f(\bm x)$ of the encoder from LIA. }
   \vspace{-0.cm}
\label{fig:avg-loss}
\end{figure}

Figure~\ref{fig:finetuning} shows the results using three different initialization methods, in which we can see that the reconstructed faces from LIA are the best.
For example, Mona Lisa's hairstyle near the right eye is correctly recovered from the latent code of LIA, whereas both the Image2StyleGAN and random sampling fail. 
Figure~\ref{fig:fine-tuning-detail} compares the reconstruction quality of the facial parts.
Table~\ref{tab:quat-finetuing-high-dimension} reports the quantitative results with those three different initialization methods using the last 1,000 images in the FFHQ dataset.
We can see that the initialization from our encoder also produces the best results.
Moreover, the average losses at each iteration during optimization are plotted in Figure~\ref{fig:avg-loss}.
For each curve, the losses are calculated on the last 1,000 images on FFHQ. It is clear that using the code from our method results in faster convergence and better reconstruction, which also demonstrates the consistent superiority of LIA.

\subsection{Disentanglement vs. Entanglement for GAN Inference}\label{exp:dis-comparison}

Besides the analysis on encoder learning , this section further shows that even for finetuning on a specific image, it will fail when optimizing in the entangled $ \bm z $-space.
As shown in section~\ref{exp:finetuning}, we have three different initialization methods when finetuning on a specific image.
Here we continue using these three initialization methods to conduct the experiments in the $ \bm z $-space and $ \bm y $-space, respectively.
Note that the difference between the encoder learning and finetuning on a specific image is that learning the encoder needs to minimize the reconstruction loss (e.g. Equation~(\ref{eq:lia-rec-loss})) on the entire dataset.
In contrast, optimization through Equation~(\ref{eq:gan-inversion2}) just needs to minimize the reconstruction loss on a specific image.
For the $\bm z$-space, the initial $\bm z_0$s for three methods are derived from $\bm y$-to-$\bm z$ mapping via the invertible network, $i.e.$ $\bm z_0 = \phi(\bm y_0)$.
Specifically, three algorithms use exactly the same initial values in the $\bm z$-space and  $\bm y$-space, as the second row shows in Figure~\ref{fig:optimization_on_yz}. 
From Figures~\ref{fig:optimization_on_yz}(c) and (d), we can see that the reconstruction is substantially improved when the reconstruction is performed in the $\bm y$-space compared to the $\bm z$-space.
These three algorithms all fail to converge at the correct minima in the $\bm z$-space.  
Table~\ref{tab:quat-different-space} shows the qualitative results on different spaces, which also indicates that the optimization in the $\bm y$-space obtains much better results than that in the $\bm z$-space.
Recall that the $ \bm y $-space is of disentanglement and the $ \bm z $-space is of entanglement, which is another evidence that the entanglement is the key factor to prevent the success of the inference for GANs.

\section{Conclusion}\label{se:conclusion}
To address the inference problem in GANs, a new generative model, named Latently Invertible Autoencoder (LIA), has been proposed to generate image samples from a probability prior and simultaneously infer accurate latent codes for a given sample.
The core idea of LIA is to embed an invertible network in an autoencoder symmetrically. Then the neural architecture can be trained with adversarial learning as two detached modules.
With the design of two-stage training, the decoder can be replaced with any GAN generator embedded with an invertible network for  image generation of diverse purposes.
The effectiveness of LIA is validated with experiments of reconstruction (inference and generation) on FFHQ and LSUN datasets. 
With the accurate inference of latent codes from LIA, future work will explore various applications based on GAN models, such as image editing, data augmentation, few-shot learning, and 3D vision. 

\textbf{Acknowledgement} The project was partially supported through the Research Grants Council (RGC) of Hong
Kong under ECS Grant No.24206219, GRF Grant No.14204521, CUHK FoE RSFS Grant.

\bibliographystyle{spbasic}
\bibliography{dlbib}

\begin{thebibliography}{62}
\providecommand{\natexlab}[1]{#1}
\providecommand{\url}[1]{{#1}}
\providecommand{\urlprefix}{URL }
\expandafter\ifx\csname urlstyle\endcsname\relax
  \providecommand{\doi}[1]{DOI~\discretionary{}{}{}#1}\else
  \providecommand{\doi}{DOI~\discretionary{}{}{}\begingroup
  \urlstyle{rm}\Url}\fi
\providecommand{\eprint}[2][]{\url{#2}}

\bibitem[{Abdal et~al.(2019)Abdal, Qin, and Wonka}]{image2stylegan}
Abdal R, Qin Y, Wonka P (2019) {Image2StyleGAN}: How to embed images into the
  {StyleGAN} latent space? In: Proceedings of International Conference on
  Computer Vision (ICCV)

\bibitem[{Antoniou et~al.(2017)Antoniou, Storkey, and
  Edwards}]{Antoniou2017augmentation}
Antoniou A, Storkey A, Edwards H (2017) Data augmentation generative
  adversarial networks. arXiv:171104340

\bibitem[{Arjovsky et~al.(2017)Arjovsky, Chintala, and Bottou}]{WGAN17}
Arjovsky M, Chintala S, Bottou L (2017) Wasserstein {GAN}. In: arXiv:1701.07875

\bibitem[{Belkin and Niyogi(2003)}]{Belkin2003}
Belkin M, Niyogi P (2003) Laplacian eigenmaps for dimensionality reduction and
  data representation. Neural Computation 15:1373--1396

\bibitem[{Berthelot et~al.(2017)Berthelot, Schumm, and Metz}]{BEGAN17}
Berthelot D, Schumm T, Metz L (2017) {BEGAN:} boundary equilibrium generative
  adversarial networks. arXiv:170310717

\bibitem[{Bowman et~al.(2015)Bowman, Vilnis, Vinyals, Dai, Jozefowicz, and
  Bengio}]{Bowman15}
Bowman SR, Vilnis L, Vinyals O, Dai AM, Jozefowicz R, Bengio S (2015)
  Generating sentences from a continuous space. In: arXiv:1511.06349

\bibitem[{Brock et~al.(2018)Brock, Donahue, and Simonyan}]{BigGAN}
Brock A, Donahue J, Simonyan K (2018) Large scale {GAN} training for high
  fidelity natural image synthesis. In: arXiv:1809.11096

\bibitem[{Creswell and Bharath(2016)}]{Creswell16}
Creswell A, Bharath AA (2016) Inverting the generator of a generative
  adversarial network. In: Advances in Neural Information Processing Systems
  (NeurIPS)

\bibitem[{Dinh et~al.(2015)Dinh, Krueger, and Bengio}]{NICE15}
Dinh L, Krueger D, Bengio Y (2015) {NICE}: Non-linear independent components
  estimation. In: International Conference on Learning Representations (ICLR)

\bibitem[{Dinh et~al.(2017)Dinh, Sohl-Dickstein, and Bengio}]{NVP17}
Dinh L, Sohl-Dickstein J, Bengio S (2017) Density estimation using {Real NVP}.
  In: International Conference on Learning Representations (ICLR)

\bibitem[{Doersch(2016)}]{VAE-tutorial16}
Doersch C (2016) Tutorial on variational autoencoders. arXiv:160605908

\bibitem[{Donahue and Simonyan(2019)}]{BigBiGAN2019}
Donahue J, Simonyan K (2019) Large scale adversarial representation learning.
  arXiv:190702544

\bibitem[{Donahue et~al.(2017)Donahue, Krahenbuhl, and Darrell}]{BiGAN2017}
Donahue J, Krahenbuhl P, Darrell T (2017) Adversarial feature learning. In:
  International Conference on Learning Representations (ICLR)

\bibitem[{Dumoulin et~al.(2017)Dumoulin, Belghazi, Poole, Mastropietro, Lamb,
  Arjovsky, and Courville}]{ALI2017}
Dumoulin V, Belghazi I, Poole B, Mastropietro O, Lamb A, Arjovsky M, Courville
  A (2017) Adversarially learned inference. In: International Conference on
  Learning Representations (ICLR)

\bibitem[{Eslami et~al.(2018)Eslami, Rezende, Besse, Viola, Morcos, Garnelo,
  Ruderman, Rusu, Danihelka, Gregor, Reichert, Buesing, Weber, Vinyals,
  Rosenbaum, Rabinowitz, King, Hillier, Botvinick, Wierstra, Kavukcuoglu, and
  Hassabis}]{sceneRendering18}
Eslami SMA, Rezende DJ, Besse F, Viola F, Morcos AS, Garnelo M, Ruderman A,
  Rusu AA, Danihelka I, Gregor K, Reichert DP, Buesing L, Weber T, Vinyals O,
  Rosenbaum D, Rabinowitz N, King H, Hillier C, Botvinick M, Wierstra D,
  Kavukcuoglu K, Hassabis D (2018) Neural scene representation and rendering.
  Science 360(6394):1204--1210

\bibitem[{Goodfellow et~al.(2014)Goodfellow, Pouget-Abadie, Mirza, Xu,
  Warde-Farley, Ozair, Courville, and Bengio}]{GAN}
Goodfellow IJ, Pouget-Abadie J, Mirza M, Xu B, Warde-Farley D, Ozair S,
  Courville A, Bengio Y (2014) Generative adversarial networks. In: Advances in
  Neural Information Processing Systems (NeurIPS)

\bibitem[{Grover et~al.(2017)Grover, Dhar, and Ermon}]{flowGAN17}
Grover A, Dhar M, Ermon S (2017) {Flow-GAN}: Combining maximum likelihood and
  adversarial learning in generative models. In: arXiv:1705.08868

\bibitem[{Gulrajani et~al.(2017)Gulrajani, Ahmed, Arjovsky, Dumoulin, and
  Courville}]{WGAN-GP17}
Gulrajani I, Ahmed F, Arjovsky M, Dumoulin V, Courville A (2017) Improved
  training of {W}asserstein {GANs}. In: arXiv:1704.00028

\bibitem[{Heljakka et~al.(2018)Heljakka, Solin, and Kannala}]{pioneerNet18}
Heljakka A, Solin A, Kannala J (2018) Pioneer networks: Progressively growing
  generative autoencoder. In: arXiv:1807.03026

\bibitem[{Huang and Belongie(2017)}]{Huang2017AdaIN}
Huang X, Belongie S (2017) Arbitrary style transfer in real-time with adaptive
  instance normalization. In: Proceedings of International Conference on
  Computer Vision (ICCV)

\bibitem[{Isola et~al.(2017)Isola, Zhu, Zhou, and Efros}]{pix2pix}
Isola P, Zhu JY, Zhou T, Efros AA (2017) Image-to-image translation with
  conditional adversarial networks. In: Proceedings of the IEEE Conference on
  Computer Vision and Pattern Recognition (CVPR)

\bibitem[{Johnson et~al.(2016)Johnson, Alahi, and Fei-Fei}]{perceptualLoss2016}
Johnson J, Alahi A, Fei-Fei L (2016) Perceptual losses for real-time style
  transfer and super-resolution. arXiv:160308155

\bibitem[{Karras et~al.(2018{\natexlab{a}})Karras, Aila, Laine, and
  Lehtinen}]{PGGAN}
Karras T, Aila T, Laine S, Lehtinen J (2018{\natexlab{a}}) Progressive growing
  of {GAN}s for improved quality, stability, and variation. In: Proceedings of
  the 6th International Conference on Learning Representations (ICLR)

\bibitem[{Karras et~al.(2018{\natexlab{b}})Karras, Laine, and
  Aila}]{StyleGAN18}
Karras T, Laine S, Aila T (2018{\natexlab{b}}) A style-based generator
  architecture for generative adversarial networks. arXiv:181204948

\bibitem[{Karras et~al.(2019)Karras, Laine, Aittala, Hellsten, Lehtinen, and
  Aila}]{StyleGAN2019}
Karras T, Laine S, Aittala M, Hellsten J, Lehtinen J, Aila T (2019) Analyzing
  and improving the image quality of {StyleGAN}. arXiv:191204958

\bibitem[{Kingma and Dhariwal(2018)}]{Glow18}
Kingma DP, Dhariwal P (2018) Glow: Generative flow with invertible 1x1
  convolutions. In: arXiv:1807.03039

\bibitem[{Kingma and Welling(2013)}]{VAE}
Kingma DP, Welling M (2013) Auto-encoding variational {B}ayes. In: Proceedings
  of the 2th International Conference on Learning Representations (ICLR)

\bibitem[{Kingma et~al.(2016)Kingma, Salimans, Jozefowicz, Chen, Sutskever, and
  Welling}]{Kingma16}
Kingma DP, Salimans T, Jozefowicz R, Chen X, Sutskever I, Welling M (2016)
  Improving variational inference with inverse autoregressive flow. In:
  arXiv:1606.04934

\bibitem[{Larsen et~al.(2016)Larsen, S{\o}nderby, Larochelle, and
  Winther}]{VAEGAN}
Larsen ABL, S{\o}nderby SK, Larochelle H, Winther O (2016) Autoencoding beyond
  pixels using a learned similarity metric. In: International Conference on
  Machine Learning (ICML), pp 1558--1566

\bibitem[{Lehtinen et~al.(2018)Lehtinen, Munkberg, Hasselgren, Laine, Karras,
  Aittala, and Aila}]{noise2noise}
Lehtinen J, Munkberg J, Hasselgren J, Laine S, Karras T, Aittala M, Aila T
  (2018) {Noise2Noise}: Learning image restoration without clean data. In:
  Proceedings of the 35th International Conference on Machine Learning (ICML)

\bibitem[{Lipton and Tripathi(2017)}]{Lipton17}
Lipton ZC, Tripathi S (2017) Precise recovery of latent vectors from generative
  adversarial networks. In: International Conference on Learning
  Representations (ICLR)

\bibitem[{Lucas et~al.(2019{\natexlab{a}})Lucas, Tucker, Grosse, and
  Norouzi}]{Lucas2019Collapse}
Lucas J, Tucker G, Grosse R, Norouzi M (2019{\natexlab{a}}) Understanding
  posterior collapse in generative latent variable models. In: Proceedings of
  International Conference on Learning Representations (ICLR)

\bibitem[{Lucas et~al.(2019{\natexlab{b}})Lucas, Shmelkov, Alahari, Schmid, and
  Verbeek}]{Lucas2019}
Lucas T, Shmelkov K, Alahari K, Schmid C, Verbeek J (2019{\natexlab{b}})
  Adversarial training of partially invertible variational autoencoders.
  arXiv:190101091

\bibitem[{Luo et~al.(2017)Luo, Xu, Tang, and Lv}]{inverseGenerator17}
Luo J, Xu Y, Tang C, Lv J (2017) Learning inverse mapping by autoencoder based
  generative adversarial nets. In: arXiv:1703.10094

\bibitem[{van~der Maaten and Hinton(2008)}]{tSNE2008}
van~der Maaten L, Hinton G (2008) Visualizing data using t-{SNE}. Journal of
  Machine Learning Research (9):2579--2605

\bibitem[{Makhani(2018)}]{Makhani18}
Makhani A (2018) Implicit autoencoders. In: arXiv:1805.09804

\bibitem[{Makhzani et~al.(2015)Makhzani, Shlens, Jaitly, Goodfellow, and
  Frey}]{Makhani15}
Makhzani A, Shlens J, Jaitly N, Goodfellow I, Frey B (2015) Adversarial
  autoencoders. In: arXiv:1511.05644

\bibitem[{Mescheder et~al.(2018)Mescheder, Geiger, and Nowozin}]{whereGAN18}
Mescheder L, Geiger A, Nowozin S (2018) Which training methods for {GANs} do
  actually converge? arXiv:180104406

\bibitem[{van~den Oord et~al.(2016)van~den Oord, Dieleman, Zen, Simonyan,
  Vinyals, Graves, Kalchbrenner, Senior, and Kavukcuoglu}]{waveNet16}
van~den Oord A, Dieleman S, Zen H, Simonyan K, Vinyals O, Graves A,
  Kalchbrenner N, Senior A, Kavukcuoglu K (2016) {WaveNet}: A generative model
  for raw audio. In: arXiv:1609.03499

\bibitem[{van~den Oord et~al.(2017)van~den Oord, Li, Babuschkin, Simonyan,
  Vinyals, Kavukcuoglu, van~den Driessche, Lockhart, Cobo, Stimberg,
  Casagrande, Grewe, Noury, Dieleman, Elsen, Kalchbrenner, Zen, Graves, King,
  Walters, Belov, and Hassabis}]{waveNet17}
van~den Oord A, Li Y, Babuschkin I, Simonyan K, Vinyals O, Kavukcuoglu K,
  van~den Driessche G, Lockhart E, Cobo LC, Stimberg F, Casagrande N, Grewe D,
  Noury S, Dieleman S, Elsen E, Kalchbrenner N, Zen H, Graves A, King H,
  Walters T, Belov D, Hassabis D (2017) Parallel {WaveNet}: Fast high-fidelity
  speech synthesis. In: arXiv:1711.10433

\bibitem[{van~den Oord et~al.(2018)van~den Oord, Li, and Vinyals}]{CPC2018}
van~den Oord A, Li Y, Vinyals O (2018) Representation learning with contrastive
  predictive coding. arXiv:180703748

\bibitem[{Pidhorskyi et~al.(2020)Pidhorskyi, Adjeroh, and Doretto}]{ALAE2020}
Pidhorskyi S, Adjeroh DA, Doretto G (2020) Adversarial latent autoencoders. In:
  Proceedings of the IEEE Conference on Computer Vision and Pattern Recognition
  (CVPR)

\bibitem[{Radford et~al.(2016)Radford, Metz, and Chintala}]{DCGAN16}
Radford A, Metz L, Chintala S (2016) Unsupervised representation learning with
  deep convolutional generative adversarial networks. In: Proceedings of the
  4th International Conference on Learning Representations (ICLR)

\bibitem[{Roweis and Saul(2000)}]{Roweis00}
Roweis ST, Saul LK (2000) Nonlinear dimensionality reduction by locally linear
  embedding. Science 290(5500):2323--2326

\bibitem[{Sankaranarayanan et~al.(2017)Sankaranarayanan, Balaji, Castillo, and
  Chellappa}]{Sankaranarayanan2017domain}
Sankaranarayanan S, Balaji Y, Castillo CD, Chellappa R (2017) Generate to
  adapt: Aligning domains using generative adversarial networks. In:
  Proceedings of the IEEE Conference on Computer Vision and Pattern Recognition
  (CVPR)

\bibitem[{Shen et~al.(2020{\natexlab{a}})Shen, Gu, Tang, and
  Zhou}]{shen2020interpreting}
Shen Y, Gu J, Tang X, Zhou B (2020{\natexlab{a}}) Interpreting the latent space
  of {GAN}s for semantic face editing. In: Proceedings of the IEEE Conference
  on Computer Vision and Pattern Recognition (CVPR)

\bibitem[{Shen et~al.(2020{\natexlab{b}})Shen, Yang, Tang, and
  Zhou}]{shen2020interfacegan}
Shen Y, Yang C, Tang X, Zhou B (2020{\natexlab{b}}) Interface{GAN}:
  Interpreting the disentangled face representation learned by {GAN}s. IEEE
  Transactions on Pattern Analysis and Machine Intelligence

\bibitem[{Simonyan and Zisserman(2014)}]{VGG}
Simonyan K, Zisserman A (2014) Very deep convolutional networks for large-scale
  image recognition. CoRR abs/1409.1556

\bibitem[{Su and Wu(2018)}]{fVAE18}
Su J, Wu G (2018) {f-VAEs}: Improve {VAEs} with conditional flows. In:
  arXiv:1809.05861

\bibitem[{Tenenbaum et~al.(2000)Tenenbaum, de~Silva, and
  Langford}]{Tenenbaum00}
Tenenbaum JB, de~Silva V, Langford JC (2000) A global geometric framework for
  nonlinear dimensionality reduction. Science 290(5500):2319--2323

\bibitem[{Ulyanov et~al.(2017)Ulyanov, Vedaldi, and Lempitsky}]{Ulyanov17}
Ulyanov D, Vedaldi A, Lempitsky V (2017) It takes (only) two: Adversarial
  generator-encoder networks. In: arXiv:1704.02304

\bibitem[{Wang et~al.(2017)Wang, Yu, Zhang, Gong, Xu, Wang, Zhang, and
  Zhang}]{IRGAN}
Wang J, Yu L, Zhang W, Gong Y, Xu Y, Wang B, Zhang P, Zhang D (2017) {IRGAN}: A
  minimax game for unifying generative and discriminative information retrieval
  models. In: Proceedings of the 40th International ACM SIGIR Conference on
  Research and Development in Information Retrieval

\bibitem[{Wu et~al.(2016)Wu, Zhang, Xue, Freeman, and Tenenbaum}]{Jiajun16}
Wu J, Zhang C, Xue T, Freeman WT, Tenenbaum JB (2016) Learning a probabilistic
  latent space of object shapes via 3{D} generative adversarial modeling. In:
  Advances in Neural Information Processing Systems (NeurIPS), pp 82--90

\bibitem[{Xiao et~al.(2019)Xiao, Yan, and Amit}]{GLF2019}
Xiao Z, Yan Q, Amit Y (2019) Generative latent flow. arXiv:190510485

\bibitem[{Yang et~al.(2021)Yang, Shen, and Zhou}]{yang2021semantic}
Yang C, Shen Y, Zhou B (2021) Semantic hierarchy emerges in deep generative
  representations for scene synthesis. International Journal of Computer Vision
  129(5):1451--1466

\bibitem[{Yu et~al.(2015)Yu, Seff, Zhang, Song, Funkhouser, and Xiao}]{LSUN15}
Yu F, Seff A, Zhang Y, Song S, Funkhouser T, Xiao J (2015) {LSUN}: Construction
  of a large-scale image dataset using deep learning with humans in the loop.
  arXiv:150603365

\bibitem[{Zhang et~al.(2018)Zhang, Isola, Efros, Shechtman, and
  Wang}]{Zhang2018PPL}
Zhang R, Isola P, Efros AA, Shechtman E, Wang O (2018) The unreasonable
  effectiveness of deep features as a perceptual metric. In: Proceedings of the
  IEEE Conference on Computer Vision and Pattern Recognition (CVPR)

\bibitem[{Zhang and Zha(2004)}]{LTSA}
Zhang Z, Zha H (2004) Principal manifolds and nonlinear dimensionality
  reduction via tangent space alignment. SIAM Journal on Scientific Computing
  26(1):313--338

\bibitem[{Zhou et~al.(2003)Zhou, Weston, Gretton, Bousquet, and
  Sch\"{o}lkopf}]{Zhou:2003}
Zhou D, Weston J, Gretton A, Bousquet O, Sch\"{o}lkopf B (2003) Ranking on data
  manifolds. In: Proceedings of the 16th International Conference on Neural
  Information Processing Systems (NeurIPS)

\bibitem[{Zhu et~al.(2018)Zhu, Liu, Cauley, Rosen, and
  Rosen}]{signal-to-image18}
Zhu B, Liu JZ, Cauley SF, Rosen BR, Rosen MS (2018) Image reconstruction by
  domain-transform manifold learning. Nature 555:487--492

\bibitem[{Zhu et~al.(2020)Zhu, Shen, Zhao, and Zhou}]{Zhu2020Indomain}
Zhu J, Shen Y, Zhao D, Zhou B (2020) In-domain {GAN} inversion for real image
  editing. In: Proceedings of European Conference on Computer Vision (ECCV)

\bibitem[{Zhu et~al.(2017)Zhu, Park, Isola, and Efros}]{cycleGAN}
Zhu JY, Park T, Isola P, Efros AA (2017) Unpaired image-to-image translation
  using cycle-consistent adversarial networks. In: International Conference on
  Computer Vision (ICCV)

\end{thebibliography}

\end{document}